\pdfoutput=1
\documentclass[11pt]{article}

\usepackage[]{EMNLP2023}

\usepackage{times}
\usepackage{latexsym}

\usepackage[T1]{fontenc}
\usepackage[utf8]{inputenc}

\usepackage{microtype}

\usepackage{inconsolata}
\usepackage{graphicx}
\usepackage{xspace}
\usepackage{multirow}
\usepackage{multicol}
\usepackage{booktabs}
\usepackage{amsmath}
\usepackage{amsfonts}
\usepackage{array}
\usepackage{url}
\usepackage{amsthm}
\usepackage{amssymb}
\usepackage{color}
\usepackage{xcolor}
\usepackage{soul}
\usepackage{subfigure}
\usepackage{listings}
\usepackage{utfsym}
\usepackage{colortbl}
\usepackage{longtable}
\usepackage{tabularray}

\definecolor{darkred}{rgb}{0.6,0.0,0.0}
\definecolor{darkgreen}{rgb}{0,0.50,0}
\definecolor{lightblue}{rgb}{0.0,0.42,0.91}
\definecolor{orange}{rgb}{0.99,0.48,0.13}
\definecolor{grass}{rgb}{0.18,0.80,0.18}
\definecolor{pink}{rgb}{0.97,0.15,0.45}
\definecolor{codegreen}{rgb}{0,0.6,0}
\definecolor{codegray}{rgb}{0.5,0.5,0.5}
\definecolor{codepurple}{rgb}{0.58,0,0.82}
\definecolor{backcolour}{rgb}{0.95,0.95,0.92}
\definecolor{LightCyan}{rgb}{0.88,1,1}

\lstdefinestyle{mystyle}{
  frame=single,
  basicstyle=\ttfamily\footnotesize,
  backgroundcolor=\color{backcolour}, commentstyle=\color{codegreen},
  commentstyle=\color{darkgreen}\slshape,
  keywordstyle=\color{blue},
  stringstyle=\color{darkred},
  numberstyle=\tiny\color{codegray},
  emphstyle=\color{pink}\underbar,
  morekeywords={Verify, Question},
  escapeinside={(*@}{@*)},
  breakatwhitespace=false,         
  breaklines=true,                 
  captionpos=b,                    
  keepspaces=true,                    
  numbersep=5pt,                  
  showspaces=false,                
  showstringspaces=false,
  showtabs=false,                  
  tabsize=2
}

\newcolumntype{L}[1]{>{\raggedright\arraybackslash}p{#1}}
\newcolumntype{C}[1]{>{\centering\arraybackslash}p{#1}}
\newcolumntype{R}[1]{>{\raggedleft\arraybackslash}p{#1}}

\newcommand{\authorspace}{,\hspace{0.3cm}}

\title{Automatically Correcting Large Language Models: \\ \textit{Surveying the landscape of diverse self-correction strategies}}

\author{
 \bf Liangming Pan\authorspace
 \bf Michael Saxon\authorspace  
 \bf Wenda Xu\authorspace \\
 \bf Deepak Nathani\authorspace
  \bf Xinyi Wang\authorspace
 \bf William Yang Wang \vspace{3mm}
 \\
 University of California, Santa Barbara \vspace{1mm}
 \\
\texttt{\{liangmingpan, saxon, wendaxu, dnathani, xinyi\_wang\}@ucsb.edu} \\
\texttt{william@cs.ucsb.edu}
}
\begin{document}
\maketitle
\begin{abstract}
Large language models (LLMs) have demonstrated remarkable performance across a wide array of NLP tasks. However, their efficacy is undermined by undesired and inconsistent behaviors, including hallucination, unfaithful reasoning, and toxic content. A promising approach to rectify these flaws is \textit{self-correction}, 
where the LLM itself is prompted or guided to fix problems in its own output.
%drawing parallels to human learning methods of trial, error, and correction. 
Techniques leveraging \textit{automated feedback}---either produced by the LLM itself or some external system---are of particular interest as they are a promising way to make LLM-based solutions more practical and deployable with minimal human feedback. 
%This paper presents a comprehensive review and discussion on this emerging area, with a focus on utilizing \textit{automated feedback} that minimizes the need for human intervention.
This paper presents a comprehensive review of this emerging class of techniques.
We analyze and taxonomize a wide array of recent work utilizing these strategies, including training-time, generation-time, and post-hoc correction. We also summarize the major applications of this strategy and conclude by discussing future directions and challenges. 

% We also summarize the major applications of this strategy, including correcting factual errors, reasoning, code generation, etc. We conclude by discussing future directions and challenges. 

% We also summarize the major applications of this strategy and conclude by discussing future directions and challenges. 

% in this rapidly evolving field. 
% We also discuss the fundamental LLM capabilities that make this ``self-correction'' possible. We conclude by summarizing key findings and implications and discussing future directions and challenges in this rapidly evolving field. 
\end{abstract}

\section{Introduction}
%%%%%% LLM is strong but still has many drawbacks
Recent years have seen striking empirical successes of large language models (LLMs), as they consistently obtain impressive results across a diverse range of NLP benchmarks~\cite{guo2023close,suzgun-etal-2023-challenging,qin2023chatgpt}, while also showcasing surprising abilities of language understanding~\cite{DBLP:Emergent_Abilities,beguš2023large}, generation~\cite{pu-demberg-2023-chatgpt,lin2023llmeval,lyu2023new}, and reasoning~\cite{wei2023chainofthought,DBLP:conf/nips/KojimaGRMI22,dasgupta2022language}. However, these models are not without their flaws. LLMs are observed to intermittently display undesired and inconsistent behaviors such as producing seemingly convincing but inaccurate ``hallucinations''~\cite{lin-etal-2022-truthfulqa,zhang2023language,min2023factscore}, conducting unfaithful reasoning~\cite{DBLP:ROSCOE,DBLP:faithful_CoT,wu2023reasoning}, generating inappropriate or harmful content~\cite{gehman-etal-2020-realtoxicityprompts,levy-etal-2021-investigating,levy-etal-2022-safetext,shaikh-etal-2023-second}, and failing to trustfully follow rules and constraints~\cite{DBLP:zhuo2023red,wang2023decodingtrust}. Such flawed behaviors hamper the trust in LLMs and pose hurdles to their real-world applications~\cite{DBLP:GPT4}. 
% producing seemingly convincing but inaccurate ``hallucinations''
% fabricating inaccurate facts (``hallucination'')
% We shouldn't cite Bubeck sparks of agi unless we're responding it, as it is a controversial and   questionable paper

%%%%%% Humans can correct errors with feedback
A prevailing strategy to rectify these undesired behaviors of LLMs is \textit{learning from feedback}, mirroring a typical human learning strategy where individuals actively refine their behaviors through a cycle of trial, error, and correction. Humans, when making mistakes, often gather feedback either from others or through self-reflection. Such feedback offers valuable insights into missteps and proposes potential avenues for improvement. With feedback, humans can adapt and modify their behavior accordingly, learning to correct their mistakes over time. 
% This capability to learn from feedback is a key component of human intelligence, contributing significantly to our ability to acquire new skills, improve performance, and solve complex problems. 
Inspired by this natural learning mechanism, extensive research~\cite{huang2022large,madaan2023selfrefine,gero2023selfverification,jiang2023selfevolve} has been undertaken to improve LLMs through the paradigm of learning from feedback. 

%%%%%% Learning from human feedback
% Leveraging \textit{human feedback} to evaluate and improve models is a popular approach. 
One popular line of research involves the use of \textit{human feedback} to evaluate and refine models, as encapsulated in the survey by ~\citet{fernandes2023bridging}. These methods typically involve direct optimization of LLMs against human feedback on their outputs~\cite{kreutzer-etal-2018-neural,glaese2022improving,DBLP:conf/nips/Ouyang0JAWMZASR22,scheurer2023training}, wherein human evaluations of output quality serve as a reward signal for improving the model performance. However, this approach has two primary drawbacks: it can be costly due to the manual labor involved, and it lacks real-time capabilities as humans cannot provide instant feedback. 

\begin{figure*}[!t]
	\centering
	\includegraphics[width=16cm]{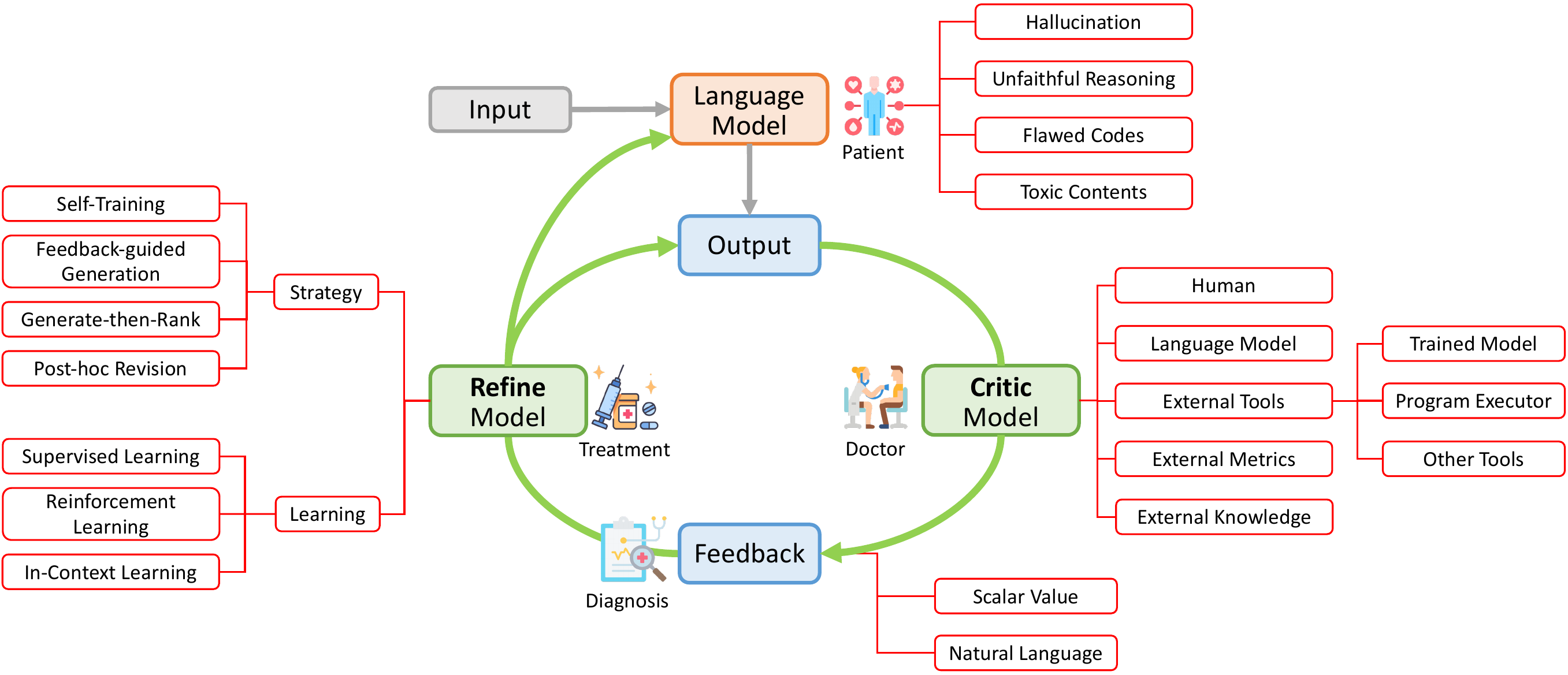}
    \caption{A conceptual framework for self-correcting LLMs with automated feedback. We identify three parties involved in the prototypical correction pipeline that are analogous to a patient, doctor, and treatment in medicine, respectively: a \textit{Language Model} produces initial output, a \textit{Critic Model} analyzes the output and provides feedback, and a \textit{Refine Model} provides treatment to either the output or the language model. We taxonomize existing works using this conceptualization along five key aspects: the \textit{problem} to be corrected, the \textit{source} and \textit{format} of the feedback, and the \textit{strategy} and \textit{learning} method of the refine model.}
    % Taxonomy of methods that leverage human-feedback, with some example representative works in the literature that fit in each category.
    \vspace{-0.2cm}
    \label{fig:general_framework}
\end{figure*}

% recent works have started to explore the possibility
To minimize the need for human intervention, another strategy is \textit{self-correcting LLMs with automated feedback}, where the model (iteratively) learns from automatically generated feedback signals to understand the consequences of its actions and adapts its behaviors. 
% The automated feedback can originate from various sources, including
The source of automated feedback can be multifaceted, spanning from the LLM itself acting as the feedback model~\cite{madaan2023selfrefine,schick2022peer}, a separately trained feedback model~\cite{yang-etal-2022-re3,paul2023refiner}, readily available external tools~\cite{Gou2023CRITICLL,chen2023teaching}, to external knowledge sources such as Wikipedia or the internet~\cite{yu2023improving,li2023selfchecker}. Different strategies have been proposed to correct LLM with automated feedback, including self-training~\cite{huang2022large,bai2022constitutional}, generate-then-rank~\cite{he2022rethinking,weng2023large}, feedback-guided decoding~\cite{DBLP:conf/emnlp/Yang0C22,xie2023decomposition}, iterative post-hoc revision~\cite{zhang2023selfedit,jiang2023selfevolve}, etc. Recently, the incorporation of such strategies has demonstrated their effectiveness across a myriad of tasks, from question answering~\cite{peng2023check} and reasoning~\cite{pan2023logiclm} to code generation~\cite{zhang2023algo} and toxicity detection~\cite{DBLP:conf/nips/LuWHJQWA022}. 

% In light of these advancements, our paper aims to provide a comprehensive overview and an insightful discussion on this fast-evolving topic. 
% In light of these advancements, our paper aims to provide a comprehensive survey of this important topic, with a primary focus on automated feedback. 

In light of these advancements, our paper aims to provide a comprehensive survey. We start by establishing the concept of \textit{self-correcting LLMs with automated feedback} and creating a taxonomy of the different methods (\S~\ref{sec:conceptualization}). We then discuss the major techniques, categorized as training-time correction (\S~\ref{sec:prehoc_methods}), generation-time correction (\S~\ref{sec:feedback_guided_methods}), and post-hoc correction (\S~\ref{sec:posthoc_methods}). 
We then summarize the major application areas of this strategy (\S~\ref{sec:applications}). Finally, we discuss key future directions (\S~\ref{sec:discussion}). 

\section{A Taxonomy for Correcting LLMs with Automated Feedback}
\label{sec:conceptualization}

%%%%%% SCOPE STATEMENT
% Under the \textit{interactive natural language processing} framework proposed by \citet{wang2023interactive}, we primarily consider collaborative correction and self-healing efforts to fall under \textit{interacting with knowledge bases} and \textit{interactive with models and tools}.

% Statement of our survey's differentiation from theirs: 
% While the things we describe can fit in their taxonomy as described above, our focus is on documenting the narrower issue of self-healing and collaborative correction rather than the broad concept of interactivity in LLMs, which additionally includes end-users

For the sake of clean exposition, we first present a conceptual framework outlining the overall process of correcting LLMs with feedback, thereby establishing the scope of this survey (\S~\ref{subsec:conceptual_framework}). 
% Following the establishment of this fundamental framework, 
We then proceed to identify five primary dimensions that serve as classification criteria for existing works: 1) What gets corrected, 2) What is the source of the feedback, 3) What is the format of the feedback, 4) When the feedback is used, and 5) How to correct the model with feedback (\S~\ref{subsec:what_is_corrected}--\S~\ref{subsec:how_to_correct}). Finally, we summarize existing works in \S~\ref{subsec:summary_works}. 

\subsection{Conceptual Framework}
\label{subsec:conceptual_framework}

We formulate the general process of correcting LLMs with automated feedback in Figure~\ref{fig:general_framework}, using an analogy of medical treatment in our daily life. Three parties are involved in this process: 

\vspace{0.15cm}

\noindent $\bullet$ \textbf{Language Model} \textit{(Patient)}. A language model $\mathcal{M}: \mathcal{X} \rightarrow \mathcal{Y}$ performs a specific task by mapping an input $x \in \mathcal{X}$ to an output text $\hat y \in \mathcal{Y}$. This formulation encompasses a wide range of NLP tasks, for example, in summarization, $x$ is a passage, $\hat y$ is the generated summary; for question-answering, $x$ is a question and $\hat y$ is the predicted answer. The initial generation $\hat y$ may be imperfect and suffer from various problems such as hallucination and incorrect reasoning. 

\vspace{0.1cm}

\noindent $\bullet$ \textbf{Critic Model} \textit{(Doctor \& Diagnosis)}. A critic model $\mathcal{C}: \mathcal{X} \times \mathcal{Y} \rightarrow \mathcal{F}$ learns to generate feedback $x, \hat y \rightarrow c$ where $\hat y \sim \mathcal{M}(x)$ is the output or partial output of the language model, and $c$ is the feedback of some format, \textit{e.g.}, scalar value, or natural language. A simple example is binary feedback of whether the output is good or bad given the input ($\mathcal{C}: \mathcal{X} \times \mathcal{Y} \rightarrow \{0, 1\}$). 

\vspace{0.1cm}

\noindent $\bullet$ \textbf{Refine Model} \textit{(Treatment)}. A refine model $\mathcal{R}: \mathcal{X} \times \mathcal{Y} \times \mathcal{F} \rightarrow \mathcal{Y}$ learns to repair an output $x, \hat y, c \rightarrow y_{new}$ based on the feedback $c$, where $y_{new}$ is the revised output. Besides repairing output, some refine models directly repair the language model $\mathcal{M}$ through fine-tuning or reinforcement learning. 

\vspace{0.15cm}

Based on the above formulation, Figure~\ref{fig:general_framework} illustrates the fundamental interaction among the language model $\mathcal{M}$, the critic model $\mathcal{C}$, and the refine model $\mathcal{R}$. However, the specific model design in existing works varies along five crucial axes, which we will elaborate on in the following sections. 

\subsection{What gets corrected?}
\label{subsec:what_is_corrected}

LLM-based natural language systems can exhibit a variety of errors. We summarize the four major types of errors that are targeted for correction in existing works through automated feedback. 
% Depending on the use case of our model, the feedback can have a variety of purposes, ranging from assessing model performance and accuracy to preventing toxicity and harmful behavior.

\vspace{0.15cm}

\noindent $\bullet$ \textbf{Hallucination.} An open challenge for LLMs is that they often hallucinate by making up facts or citing sources that do not exist~\cite{li2023halueval,zhang2023language}. These hallucinated contents are often quite plausible-sounding, making it difficult even for humans to detect~\cite{DBLP:conf/acl/ClarkASHGS20}. To address this, several studies have proposed the collection of automated feedback on potential factual inaccuracies by cross-referencing the output generated by the model with credible knowledge sources. The gathered feedback can then be utilized by a subsequent refinement model to correct hallucinations~\cite{gao2023rarr,peng2023check}. 

\vspace{0.15cm}

% Moreover, the integration of external tools for the provision of credible information sources can help alleviate the hallucination issue [72, 469, 471]. 
% generate plausible-sounding but potentially incorrect or misleading information. Automated feedback can help correct hallucinations by comparing the generated output with a known source of truth. 

% Reference: Survey of LLMs pp 44-45
\noindent $\bullet$ \textbf{Unfaithful Reasoning.} LLMs have exhibited a strong ability in solving complex reasoning tasks with improved reasoning strategies, such as Chain-of-Thought prompting~\cite{wei2023chainofthought}. However, recent studies~\cite{DBLP:ROSCOE,DBLP:STREET,DBLP:faithful_CoT} found that LLMs occasionally make \textit{unfaithful} reasoning, \textit{i.e.}, the derived conclusion does not follow the previously generated reasoning chain. To address this, existing works have proposed the use of automated feedback from external tools or models for guiding the reasoning process~\cite{xie2023decomposition,yao2023tree}, verifying the reasoning process and rectifying errors~\cite{he2022rethinking,pan2023logiclm}, or fine-tuning LLMs with process-based feedback~\cite{huang2022large,lightman2023lets}.

\vspace{0.15cm}

\noindent $\bullet$ \textbf{Toxic, Biased, and Harmful Contents.}
% Reference: Harmlessness in "bridge the gap" paper
% Reference: pp57 in Survey of LLMs
% As the major approach to averting these issues, reinforcement learning from human feedback (RLHF) [61, 100] has been widely used by incorporating humans in the training loop for developing well-aligned LLMs. 
% It is necessary to align LLMs with human values, e.g., helpful, honest, and harmless. For this purpose, InstructGPT designs an effective tuning approach that enables LLMs to follow the expected instructions, which utilizes the technique of reinforcement learning with human feedback. 
% (survey of LLMs) potentially risky response within some specific context [46]. 
LLMs have been observed to occasionally generate content that is toxic, biased, or harmful due to biases present in the training data~\cite{shaikh-etal-2023-second}. To rectify this, reinforcement learning from human feedback (RLHF)~\cite{DBLP:conf/nips/Ouyang0JAWMZASR22,bai2022training} has been extensively employed to train LLMs to align more closely with human values, such as being helpful, honest, and harmless. However, RLHF is heavily dependent on high-quality human feedback, the collection of which can be resource-intensive. To alleviate this, recent works~\cite{DBLP:conf/nips/LuWHJQWA022,Gou2023CRITICLL} have also explored collecting automated feedback to identify and correct potentially harmful outputs. 

\vspace{0.15cm}

% LLMs are often used to generate code, but the generated code can sometimes be flawed or incorrect. Automated feedback can correct this by comparing the generated code with a known correct version or by using a code-checking tool to identify and correct errors. 
\noindent $\bullet$ \textbf{Flawed Code.} Besides generating natural language text, LLMs also show strong abilities to generate computer programs (\textit{i.e.}, code)~\cite{DBLP:conf/iclr/ChenZNZLLC23}. However, the generated code can sometimes be flawed or incorrect. To fix this, the approach of learning from automated feedback has been extensively applied in code generation~\cite{chen2023teaching,olausson2023demystifying}, largely facilitated by the ease of obtaining such feedback through the execution of generated code with the corresponding compilers or interpreters. 

% Unlike natural language generation, as the generated code can be directly checked by execution with corresponding compilers or interpreters, existing work mostly evaluates

% To make the LLM more faithful in reasoning; more factual in knowledge (reducing hallucinations; Improving correctness in objective outputs; Improving subjective style according to self-assessed attributes?

% model editing is not in our scope. 
\subsection{What is the \textit{source} of the feedback?}
\label{subsec:source_of_feedback}
% Reference: Section 6.3.3; Survey of LLM
% Reference: Section 7; Bridge the Gap
% Reference: Section 3.1.2; 3.1.3 Reasoning with Language Model Prompting: A Survey

Feedback can be broadly divided into two categories: \textit{human feedback} and \textit{automated feedback}. \citet{fernandes2023bridging} provided a survey on integrating human feedback for language generation. In our survey, we focus on the emerging research area of automated feedback, which explores the possibility of LLMs to self-correct without constant human intervention. Automated feedback typically originates from two sources, distinguished by their relationship with the LLM: \textit{self-feedback} (\textit{i.e.}, the feedback originates from the LLM itself) and \textit{external feedback} (\textit{i.e.}, the feedback is derived from external models, tools, or knowledge sources). 

\vspace{0.15cm}

\noindent $\bullet$ \textbf{Self-Feedback.}
The LLM itself can be utilized as a feedback provider. One straightforward way is to directly evaluate the quality of the generated outputs through prompting and subsequently use this feedback to refine the results~\cite{madaan2023selfrefine,shinn2023reflexion}. This process can be iterative, with the model continually refining its output until it meets a certain standard. This continuous self-improvement strategy has been found particularly useful by numerous studies~\cite{selfee2023,DBLP:conf/acl/YanSTWYY23}, especially in scenarios where external feedback is unavailable or limited. 

% Numerous studies have found that LLMs have a remarkable ability for self-analysis and self-improvement. 
% Self-feedback can be particularly useful in scenarios where external feedback is unavailable or limited. 
% In this scenario, the model engages in a continuous self-improvement process, learning from its evaluations and refining its capabilities accordingly. 

\vspace{0.15cm}

\noindent $\bullet$ \textbf{External Feedback.} Feedback can originate from sources external to the LLM, typically including 1) other trained models~\cite{yang-etal-2022-re3,lightman2023lets}, 2) external tools~\cite{Gou2023CRITICLL,charalambous2023new}, 3) external knowledge sources~\cite{gao2023rarr,yu2023improving}, and 4) external evaluation metrics~\cite{DBLP:conf/emnlp/JungQWBB0C22,DBLP:conf/iclr/WelleckLWBSK023}. External feedback provides a valuable outside perspective which is particularly useful in identifying errors that the LLM might not recognize on its own. For example, code interpreters are widely used in programming tasks to provide real-time error messages; while external knowledge sources can be utilized to verify the factual accuracy of the LLM's output. 

\subsection{What is the \textit{format} of the feedback?}
\label{subsec:format_of_feedback}

% Choosing the format of the feedback needs to consider the expressivity of the feedback, the ease of its collection, and how we can use it to improve systems~\cite{fernandes2023bridging}. The automated feedback used in existing works are typically in the form of a \textit{scalar value} signal, or in \textit{natural language}. 

The selection of feedback format requires the consideration of its expressivity, the ease of its collection, and its potential to improve systems~\cite{fernandes2023bridging}. In existing works, automated feedback is typically in the form of a \textit{scalar value} signal or in \textit{natural language}.

\vspace{0.15cm}

\noindent $\bullet$ \textbf{Scalar Value Feedback.}
In this scenario, the critic model maps the input and output to a single score ($\mathcal{C}: \mathcal{X} \times \mathcal{Y} \rightarrow \mathcal{N} \subseteq \mathbb{R}$). Scalar value feedback can be easily integrated into the training/decoding process of LLMs. For example, Self-Verification~\cite{weng2023large} ranks candidate outputs to find the optimal one based on the real-value feedback score assigned by the critic model to each candidate. Similarly, \citet{xie2023decomposition} use real-value feedback for each intermediate reasoning step to guide the model in performing a stochastic beam search for the optimal solution. However, despite its flexibility, scalar value feedback is often less informative to capture detailed information necessary for model correction. 

\vspace{0.15cm}

\noindent $\bullet$ \textbf{Natural Language Feedback.}
Natural language feedback offers greater expressivity than scalar value feedback, providing richer information that can highlight the shortcomings of the current output or suggest specific improvements. This form of feedback is particularly crucial for certain applications, such as text editing and code generation. For text editing, PEER~\cite{schick2022peer} trains an LLM to generate detailed suggestions for edits to the initial generated text, such as ``remove unsourced claim'' or ``rewrote the guacamole question for clarity''. 
For code generation, Self-Debug~\cite{chen2023teaching} uses LLMs to generate explanations for the produced code and utilize both the explanation and the execution results as feedback to enhance coding solutions. 

% cases like moral correction, NL feedback is more suitable since scalar feedback is quite subjective and uninterpretable. 
% The choice of format also has implications in the difficulty for humans to give feedback, its consistency/agreement, and the level of rationality of said feedback (Ghosal et al., 2023). 

% \tabincell{c}{QA, Program Synthesis, \\ Toxicity Reduction}
\begin{table*}[!t]
\centering
\resizebox{\textwidth}{!}{
   \begin{tabular}% 
   {lccccc}
    \toprule
        \multirow{2}{*}{\textbf{Method}} & \multicolumn{2}{c}{\textbf{Feedback}} & \multicolumn{2}{c}{\textbf{Model Refinement}} & \multirow{2}{*}{\textbf{Application}} \\ \cline{2-5}
        & \textbf{Source} & \textbf{Format} & \textbf{Strategy} & \textbf{Learning} & \\ \midrule
        \rowcolor{LightCyan} \multicolumn{6}{c}{\textbf{Training-Time Correction}} \\ \midrule
        % RLHF
        RLHF~\cite{DBLP:conf/nips/Ouyang0JAWMZASR22} & Reward Model & Scalar & RLHF & RL & Multiple Tasks \\
        % Fine-grained RLHF
        Fine-Grained RLHF~\cite{wu2023finegrained} & Reward Model & Scalar & RLHF & RL & Detoxification, Long-form QA \\
        % HH-RLHF
        HH-RLHF~\cite{bai2022training} & Reward Model & Scalar & RLHF & SL \& RL & Helpfulness, Harmlessness \\
        % Moral Self-Correction
        Moral RLHF~\cite{ganguli2023capacity} & Reward Model & Scalar & RLHF & RL & Moral Correction \\
        % Sparrow
        Sparrow~\cite{glaese2022improving} & Reward Model & NL & RLHF & SL \& RL & Dialogue \\
        % ILF
        ILF~\cite{scheurer2023training} & Human Feedback & NL & Fine-tuning & SL & Summarization \\
        % ILF (Code)
        ILF-Code~\cite{chen2023improving} & Human Feedback & NL & Fine-tuning & SL & Code Generation \\
        % SLT
        SLT~\cite{yuan2023systemlevel} & Human Feedback & NL & Fine-tuning & SL & Response Generation \\
        % Continually Improving Extractive QA via Human Feedback
        \citet{gao2023continually} & Human Feedback & NL & Fine-tuning & SL \& RL & Extractive QA \\
        % Chain-of-Hindsight
        Chain-of-Hindsight~\cite{liu2023chain} & Human Feedback & NL & Fine-tuning & SL & Multiple Tasks \\
        % Quark
        Quark~\cite{DBLP:conf/nips/LuWHJQWA022} & External Metrics & Scalar & Fine-tuning & RL & Toxicity, Repetition, Sentiment \\
        % traditional method
        % Minimum Risk Training~\cite{shen-etal-2016-minimum} & External Metrics & Scalar & Fine-tuning & SL & \usym{2717} & Machine Translation \\
        % SimCLS
        SimCLS~\cite{liu-liu-2021-simcls} & External Metrics & Scalar & Fine-tuning & SL & Summarization \\
        % BERTTune
        BERTTune~\cite{unanue2021berttune} & External Metrics & Scalar & Fine-tuning & RL & Machine Translation \\
        % STaR
        STaR~\cite{zelikman2022star} & Language Model & NL & Self-Training & SL & QA, Reasoning \\
        % Self-Instruct
        Self-Instruct~\cite{DBLP:conf/acl/WangKMLSKH23} & Language Model & NL & Self-Training & SL & Multiple Tasks \\
        % Constitutional AI
        RLAIF~\cite{bai2022constitutional} & Language Model & NL & Self-Training & SL \& RL & Dialogue \\
        % SIRLC
        SIRLC~\cite{pang2023language} & Language Model & NL & Self-Training & RL & Reasoning, Translation, Summary \\
        % Self-Improve
        Self-Improve~\cite{huang2022large} & Language Model & NL & Self-Training & SL & QA, Reasoning, NLI \\
        % AlpacaFarm
        AlpacaFarm~\cite{dubois2023alpacafarm} & Language Model & NL & Self-Training & SL \& RL & None (Intrinsic Evaluation) \\
        % ReST
        ReST~\cite{gulcehre2023reinforced} & Language Model & NL & Self-Training & RL & Machine Translation \\
        %%%%%%
        \midrule
        \rowcolor{LightCyan} \multicolumn{6}{c}{\textbf{Generation-Time Correction}} \\ 
        \midrule
        %%%%%%
        % Self-Verification
        Self-Verification~\cite{weng2023large} & Language Model & Scalar & Re-Ranking & ICL & Arithmetic Reasoning \\ 
        % CodeT
        CodeT~\cite{DBLP:conf/iclr/ChenZNZLLC23} & Program Executor & Scalar & Re-Ranking & ICL & Code Generation \\
        % LEVER
        LEVER~\cite{ni2023lever} & Program Executor & Scalar & Re-Ranking & SL & Table QA, Math QA, Program \\
        % Rethinking
        RR~\cite{he2022rethinking} & External Knowledge & Scalar & Re-Ranking & --- & Reasoning \\
        % InstructScore
        InstructScore~\cite{xu2023instructscore} & Language Model & NL & Re-Ranking & SL & Generation Evaluation \\
        % Minimum bayes risk decoding
        MBR Decoding~\cite{freitag-etal-2022-high} & External Metrics & Scalar & Re-Ranking & SL & Machine Translation \\
        % DIVERSE
        DIVERSE~\cite{li-etal-2023-making} & Trained Model & Scalar & Re-Ranking & SL & Arithmetic Reasoning \\
        % TeachMe
        TeachMe~\cite{DBLP:conf/emnlp/MishraTC22} & Human Feedback & NL & Feedback-guided & SL & Proof Generation \\
        % Let's verify step-by-step
        PRM~\cite{lightman2023lets} & Reward Model & Scalar & Feedback-guided & SL & Arithmetic Reasoning \\
        % DiffusionLM
        DiffusionLM~\cite{Li-2022-DiffusionLM} & Trained Model & Scalar & Feedback-guided & SL & Controlled Text Generation \\
        % PPLM
        PPLM~\cite{Dathathri2020Plug} & Trained Model & Scalar & Feedback-guided & SL & Controlled Text Generation \\
        % Fudge
        Fudge~\cite{yang-klein-2021-fudge} & Trained Model & Scalar & Feedback-guided & SL & Controlled Text Generation \\
        % Entailer
        Entailer~\cite{DBLP:conf/emnlp/TafjordMC22} & Trained Model & Scalar & Feedback-guided & SL & Proof Generation \\
        % NLProofS
        NLProofS~\cite{DBLP:conf/emnlp/Yang0C22} & Trained Model & Scalar & Feedback-guided & SL & Proof Generation \\
        % GRACE
        GRACE~\cite{KhalifaGRACE} & Trained Model & Scalar & Feedback-guided & SL & Arithmetic Reasoning \\
        % CoRe
        CoRe~\cite{DBLP:conf/acl/ZhuWZZ0GZY23} & Trained Model & Scalar & Feedback-guided & SL & Arithmetic Reasoning \\
        % Detection and Mitigation
        \citet{varshney2023stitch} & External Knowledge & NL & Feedback-guided & ICL & Hallucination Detection \\
        % MemPrompt
        MemPrompt~\cite{madaan-etal-2022-memory} & External Knowledge & NL & Feedback-guided & ICL &  Lexical and Ethical Reasoning \\
        % Maieutic Prompting
        Maieutic Prompting~\cite{DBLP:conf/emnlp/JungQWBB0C22} & External Metrics & Scalar & Feedback-guided & ICL & Commonsense Reasoning \\
        % Selection-Inference
        SI~\cite{DBLP:SelectionInference} & Language Model & Scalar & Feedback-guided & ICL & Proof Generation \\
        % RAP
        RAP~\cite{hao2023reasoning} & Language Model & Scalar & Feedback-guided & ICL & Planning, Reasoning \\
        % Self-Evaluation Guided Decoding
        SelfEval-Decoding~\cite{xie2023decomposition} & Language Model & Scalar & Feedback-guided & ICL & Arithmetic / Symbolic Reasoning \\
        % Self-Check
        SelfCheck~\cite{miao2023selfcheck} & Language Model & NL & Feedback-guided & ICL & Arithmetic Reasoning \\
        % Tree-of-Thought
        Tree of Thoughts~\cite{yao2023tree} & Language Model & NL / Scalar & Feedback-guided & ICL & Games, Writing \\
        \bottomrule
   \end{tabular}
} 
    \caption{Representative works on \textbf{Training-time Correction} and \textbf{Generation-Time Correction}. We summarize the key features for each work: the \textit{source} and \textit{format} of the feedback, the \textit{strategy} and \textit{learning} method of the refinement process, and the \textit{application} of the method.}
    \label{tbl:summary_works}
    \vspace{-0.4cm}
\end{table*}

\begin{table*}[!t]
\centering
\resizebox{\textwidth}{!}{
   \begin{tabular}% 
   {lcccccc}
    \toprule
        \multirow{2}{*}{\textbf{Method}} & \multicolumn{2}{c}{\textbf{Feedback}} & \multicolumn{3}{c}{\textbf{Model Refinement}} & \multirow{2}{*}{\textbf{Application}} \\ \cline{2-6}
        & \textbf{Source} & \textbf{Format} & \textbf{Strategy} & \textbf{Learning} & \textbf{Iter.} & \\ \midrule
        %%%%%%
        \rowcolor{LightCyan} \multicolumn{7}{c}{\textbf{Post-hoc Correction}} \\ 
        \midrule
        %%%%%%
        % Self-Refine
        Self-Refine~\cite{madaan2023selfrefine} & Language Model & NL & Self-Refine & ICL & \usym{2713} & Multiple Tasks \\
        % PEER
        % PEER~\cite{schick2022peer} & Language Model & NL & Post-hoc & SL & \usym{2713} & Text Editing \\
        % Clinical information extraction
        Clinical SV~\cite{gero2023selfverification} & Language Model & NL & Self-Refine & ICL & \usym{2717} & Information Extraction \\
        % Reflexion
        % \tabincell{c}{Decision-making, QA, \\ Code Generation}
        Reflexion~\cite{shinn2023reflexion} & Language Model & NL & Self-Refine & RL & \usym{2713} & QA, Code Generation \\
        % Iterative Improvement
        IterRefinement~\cite{Chen2023IterativeTR} & Language Model & NL & Self-Refine & ICL & \usym{2713} & Machine Translation \\
        % Auto-Post-Editing
        Auto-Post-Editing~\cite{raunak2023leveraging}  & Language Model & NL & Self-Refine & ICL & \usym{2717} & Machine Translation \\
        % RCI
        RCI~\cite{kim2023language} & Language Model & NL & Self-Refine & ICL & \usym{2713} & Computer Tasks \\
        % Selfee
        SelFee~\cite{selfee2023} & Language Model & NL & Self-Refine & SL & \usym{2713} & Dialogue \\
        % SelfCheckGPT
        SelfCheckGPT~\cite{manakul2023selfcheckgpt} & Language Model & NL & Self-Refine & ICL & \usym{2717} & Hallucination Detection \\
        % LLM Self Defense
        LLM Self Defense~\cite{helbling2023llm} & Language Model & NL & Self-Refine & ICL & \usym{2717} & Harmful Text Correction \\
        % Re3
        Re$^3$~\cite{yang-etal-2022-re3} & Trained Model & Scalar & External Feedback & SL \& ICL & \usym{2713} & Story Generation \\
        % CodeRL
        CodeRL~\cite{le2022coderl} & Trained Model & Scalar & External Feedback & RL & \usym{2717} & Code Generation \\
        % FLIRT
        FLIRT~\cite{mehrabi2023flirt} & Trained Model & Scalar & External Feedback & ICL & \usym{2713} & Adversarial Prompt Generation \\
        % REFINER
        REFINER~\cite{paul2023refiner} & Trained Model & NL & External Feedback & SL \& ICL & \usym{2713} & Reasoning, Moral Story \\
        % RL4F
        RL4F~\cite{DBLP:conf/acl/AkyurekAKCWT23} & Trained Model & NL & External Feedback & SL \& RL & \usym{2713} & Planning, Summarization \\
        % Feedback for Interactive Semantic Parsing
        \citet{DBLP:conf/acl/YanSTWYY23} & Trained Model & NL & External Feedback & SL & \usym{2713} & Semantic Parsing \\
        % Baldur
        Baldur~\cite{first2023baldur} & Trained Model & NL & External Feedback & ICL & \usym{2713} & Proof Generation \\
        % CRITIC
        CRITIC~\cite{Gou2023CRITICLL} & External Tools & NL & External Feedback & ICL & \usym{2713} & QA, Program, Toxicity \\ 
        % FacTool
        FacTool~\cite{chern2023factool} & External Tools & NL & External Feedback & ICL & \usym{2713} & QA, Reasoning, Generation \\
        % RARR
        RARR~\cite{gao2023rarr} & External Knowledge & NL & External Feedback & ICL & \usym{2717} & Open-Domain QA \\ 
        % LLM-Augmenter
        LLM-Augmenter~\cite{peng2023check} & External Knowledge & NL & External Feedback & RL & \usym{2713} & Open-Domain QA \\
        % Self-Checker
        Self-Checker~\cite{li2023selfchecker} & External Knowledge & NL & External Feedback & ICL & \usym{2717} & Fact-Checking \\
        % REFEED
        REFEED~\cite{yu2023improving} & External Knowledge & NL & External Feedback & ICL & \usym{2717} & QA, Dialogue \\ 
        % Demystifying
        \citet{olausson2023demystifying} & Program Executor & NL & External Feedback & ICL & \usym{2713} & Code Generation \\
        % Self-Edit
        Self-Edit~\cite{zhang2023selfedit} & Program Executor & NL & External Feedback & ICL & \usym{2713} & Code Generation \\
        % Self-debug
        Self-Debug~\cite{chen2023teaching} & Program Executor & NL & External Feedback & ICL & \usym{2713} & Code Generation \\
        % Self-Evolve
        Self-Evolve~\cite{jiang2023selfevolve} & Program Executor & NL & External Feedback & ICL & \usym{2713} & Code Generation \\
        % Logic-LM
        Logic-LM~\cite{pan2023logiclm} & Symbolic Solver & NL & External Feedback & ICL & \usym{2713} & Logical Reasoning \\
        % Self-Critique
        Self-Critique~\cite{Saunders2022SelfcritiquingMF} & LLMs + Human & NL & External Feedback & SL & \usym{2717} & Summarization \\
        % ALGO
        ALGO~\cite{zhang2023algo} & Oracle Verifier & Scalar & External Feedback & ICL & \usym{2713} & Code Generation \\
        % Self-Healing Software
        \citet{charalambous2023new} & BMC Tool & NL & External Feedback & ICL & \usym{2717} & Software Verification \\
        % Self-Correct
        Self-Correction~\cite{DBLP:conf/iclr/WelleckLWBSK023} & External Metrics & NL / Scalar & External Feedback & SL & \usym{2713} & Reasoning, Generation, Toxicity \\
        % Multiagent Debate
        Multiagent Debate~\cite{du2023improving} & Language Model & NL & Model Debate & ICL & \usym{2713} & Reasoning, Factuality \\
        % LM vs LM
        LM vs LM~\cite{cohen2023lm} & Language Model & NL & Model Debate & ICL & \usym{2713} & Factual Error Detection \\
        % ICL-AIF
        ICL-AIF~\cite{fu2023improving} & Language Model & NL & Model Debate & ICL & \usym{2713} & Bargaining Game \\
        % PRD
        PRD~\cite{li2023prd} & Language Model & NL & Model Debate & ICL & \usym{2713} & Open-ended QA \\
        \bottomrule
   \end{tabular}
} 
    \caption{Representative works on \textbf{Post-hoc Correction}. We summarize the key features for each work: the \textit{source} and \textit{format} of the feedback, the \textit{strategy} and \textit{learning} method of the refinement process, whether the refinement is iterative \textit{(Iter.)}, and the \textit{application} of the method.}
    \label{tbl:summary_works_2}
\end{table*}

% \subsection{When the feedback is used for correction?}
\subsection{\textit{When} to correct the model with feedback?}
\label{subsec:when_feedback}

Depending on the timing of using automated feedback to correct the model, existing works can be divided into three major categories. 

\vspace{0.15cm}

\noindent $\bullet$ \textbf{Training-time Correction.}
The ideal scenario is to rectify a flawed model during training, prior to its deployment for use. Once feedback has been collected, it is directly used to optimize the model parameters. Human feedback is typically used for training-time correction, as exemplified by the widely adopted RLHF approach~\cite{DBLP:conf/nips/Ouyang0JAWMZASR22}. For leveraging automated feedback, a common strategy is \textit{self-training}~\cite{huang2022large}, where the model is trained with its own generated high-quality output filtered out by the critic model. 
% \citet{korbak2023pretraining} employs conditional training \cite{Keskar2019CTRLAC} to tag the undesirable content during pretraining and controls undesirable token generation at inference. 
While training-time correction is a pre-hoc strategy that addresses problems during training, its practical application may be hindered by three issues: 1) the infeasibility of fine-tuning gaint closed-source LLMs, such as GPT-4~\cite{DBLP:GPT4}, 2) the potential unavailability of feedback during model training, and 3) the requirement for the feedback to be ``optimizable'', \textit{e.g.}, a numerical score severing as the basis for model optimization. 
% that can be used as a basis for model optimization. 
% generates output, receives automated feedback on this output, and then utilizes this feedback to update its parameters until it achieves certain quality standards. 

\vspace{0.15cm}

\noindent $\bullet$ \textbf{Generation-time Correction.} This strategy utilizes automated feedback to guide the language model \textit{during} generation, which allows the model to correct errors in its outputs as it is being generated. For example, for proof generation, several works utilize the automated feedback of the intermediate reasoning steps to guide the model to recover from incorrect generation and search for the optimal solution in a more efficient way~\cite{DBLP:conf/emnlp/Yang0C22,lightman2023lets}. 

\vspace{0.15cm}

\noindent $\bullet$ \textbf{Post-hoc Correction.} 
Finally, post-hoc correction involves refining the model output \textit{after it has been generated}, without updating the model parameters. This typically involves an iterative process of generating output, receiving feedback, and refining output. Post-hoc correction provides more flexibility than the previous two strategies as it does not require training the LLM or accessing its parameters. Furthermore, post-hoc correction enhances explainability as it facilitates the incorporation of more informative natural language feedback. This allows for a more transparent visualization and interpretation of the self-correction process. 

% Iterative Output Refinement: This method employs human feedback to refine the model’s output iteratively. Users can provide feedback on intermediate responses, enabling the model to adjust its output until it meets the user’s satisfaction. This process allows the model to better understand user preferences and produce more suitable outcomes (Reid and Neubig, 2022; Saunders et al., 2022; Schick et al., 2022; Nijkamp et al., 2022). Feedback can also be provided on model attributes such as the decoding strategy (Passali et al., 2021), rather than directly on its outputs. 

\subsection{\textit{How} to correct the model with feedback?}
\label{subsec:how_to_correct}

Various concrete strategies have been proposed to correct LLMs with automated feedback, which are tailored to the different dimensions we mentioned in previous sections. For example, \textit{self-training} is often used for training-time correction. \textit{Generate-then-rank} often comes with scalar value feedback. \textit{Self-refine} is the strategy that uses the same LLM as both the critic model and the refine model. We will cover the landscape of self-correction strategies through Section~\ref{sec:prehoc_methods} to Section~\ref{sec:posthoc_methods}. 
% We will cover typical existing strategies for model correction through Section~\ref{sec:prehoc_methods} to Section~\ref{sec:posthoc_methods}. 
% is the strategy that 

\subsection{Summary of existing works}
\label{subsec:summary_works}

Building upon the taxonomy established in the preceding sections, we collate existing works on correcting LLMs with (automated) feedback in Table~\ref{tbl:summary_works} and Table~\ref{tbl:summary_works_2}. We have two major selection criteria for a work to be included in this survey:

\vspace{0.1cm}

1. \textbf{Automated Feedback}: Explicit feedback is involved to assess the quality of the model output. We focus on automated feedback that originates from external models, metrics, knowledge, etc. However, we will cover some representative works of human feedback for completeness. 

2. \textbf{Model Refinement}: The feedback should act as a directive to enhance the LLM, either by: 1) updating model parameters, or 2) altering the model's output during or post the generation process. 

\vspace{0.1cm}

These works are categorized based on the three strategies introduced in Section~\ref{subsec:when_feedback}. We also summarize key features of each work, including: 1) the source of feedback, 2) the format of feedback, 3) the strategy and learning method employed for the refinement, 4) whether the refinement process is iterative, and 5) the application of the method. Subsequently, we will delve into a detailed review of each method type, encompassing \textit{Training-Time Correction} (\S~\ref{sec:prehoc_methods}), \textit{Generation-Time Correction} (\S~\ref{sec:feedback_guided_methods}), and \textit{Post-hoc Correction} (\S~\ref{sec:posthoc_methods}).

\begin{figure*}[!t]
    \centering
    \hypertarget{prehoc_methods}{}
        \includegraphics[width=16cm]{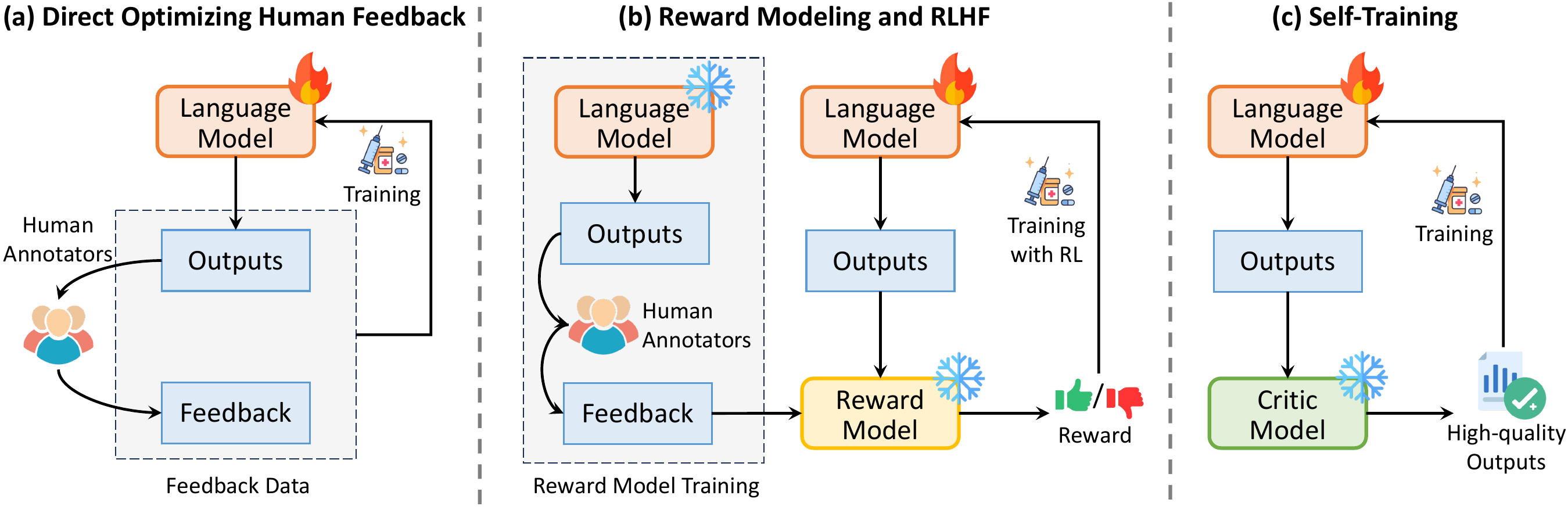}
    \caption{Three typical strategies of \textit{training-time correction}: directly optimization with human feedback (a), training a reward model that approximates human feedback (b), and self-training with automated feedback (c).}
    \label{fig:prehoc_methods}
\end{figure*}

% \section{Optimizing LLMs with Feedback}
\section{Training-Time Correction}
\label{sec:prehoc_methods}
% Reference: Sec 5.2 Alignment Tuning in Survey of Large Language Model

% In this section, we introduce the methods that correct model behavior at training time. In Figure~\ref{fig:prehoc_methods}, we list three typical strategies of training-time correction, which leverages the human feedback (a), the reward model that approximates human feedback (b), and the automated feedback (c), respectively, to update the model parameters at training time. 

In this section, we delve into methodologies that rectify model behavior during the training phase. As depicted in Figure~\ref{fig:prehoc_methods}, we identify three typical strategies for training-time correction. Each strategy utilizes different forms of feedback to modify the model parameters during training: human feedback (a), a reward model that approximates human feedback (b), and automated feedback (c). 

\subsection{Learning from Human Feedback}
\label{subsec:learn_from_human_feedback}
% \subsection{Human Value Alignment}
% \subsection{Optimizing with Human Feedback}

The next-word prediction objective of LLM pre-training is not inherently designed to encapsulate human values or preferences. This misalignment can lead to unintended consequences, such as the generation of harmful, misleading, or biased content~\cite{kenton2021alignment}. To mitigate these issues, many research efforts have explored the integration of human feedback to better align LLMs with human values and expectations. 
% , which are extensively reviewed by~\citet{wang2023aligning} and ~\citet{fernandes2023bridging}. 
\citet{wang2023aligning} and~\citet{fernandes2023bridging} extensively reviewed this research area. 
Our survey, however, focuses on automated feedback, thus we will only touch upon representative works in this direction.

% selectively. that either directly optimize LLMs with human feedback or train reward models with human feedback. 

% An extensive overview of aligning LLMs with human values is available in~\citet{wang2023aligning} and ~\citet{fernandes2023bridging}. Our survey, however, focuses on automated feedback rather than human feedback, thus we will only selectively touch upon works that either directly optimize LLMs with human feedback or train reward models with human feedback. 

% employ either \textit{fine-tuning} or \textit{reinforcement learning} to optimize LLMs using human feedback. 
% in this line for completeness

% LLMs are then fine-tuned on the outputs with positively-labeled feedback to better align with human preferences. 
% \paragraph{Fine-tuning with Human Feedback.}
\paragraph{Direct Optimization with Human Feedback.}
\label{sec:direct_optimization}
In an ideal scenario, we would directly leverage human feedback to optimize the model parameters. Typically, this approach follows the framework depicted in Figure~\hyperlink{prehoc_methods}{\ref{fig:prehoc_methods}(a)}: 1) Candidate outputs are generated by LLMs, 2) Humans provide feedback or refinements on these outputs, and 3) LLMs are then directly optimized on the collected (outputs, feedback) to better align with human preferences. A simple strategy is to fine-tune the model on the outputs with positively-labeled feedback. For example, \textit{Sparrow}~\cite{glaese2022improving} fine-tunes LLMs on the collected dialogues rated as preferred and rule compliant (concerning correctness, harmfulness, and helpfulness), according to humans. Similarly, \citet{scheurer2023training} utilizes an LLM to generate multiple refinements of the original output based on human feedback, and then the best refinement is picked up to finetune the original LLM. A similar idea is adopted to fine-tune the code generation model~\cite{chen2023improving}. First, human annotators write natural language feedback for incorrect codes. A refinement model then utilizes this feedback to correct the code. Finally, the refined code is subsequently employed to fine-tune the code-generating LLM. However, only utilizing positive data (human-refined or positive-rated data) for fine-tuning may constrain the model's ability to identify and correct negative attributes or errors. To address this, \textit{Chain-of-Hindsight}~\cite{liu2023chain} fine-tunes the LLM on model outputs paired with both positive and negative feedback. Beyond fine-tuning, other optimization methods are explored as well. For example, \citet{gao2023continually} utilizes human feedback as the reward signal and optimizes the model with contextual bandit learning. 
% Similarly, in Imitation learning from Language Feedback (ILF)~\cite{scheurer2023training}, 

% \paragraph{Reinforcement Learning from Human Feedback.}
\paragraph{Reward Modeling and RLHF.} 
\label{sec:reward_modeling}
Employing human feedback directly to rectify model behavior may not always be practical. The collection of human feedback can be both labor-intensive and time-consuming. An efficient alternative is to train a \textit{reward model} that emulates human feedback. Once trained, this reward model can provide consistent, real-time feedback for every model output, thereby circumventing the need for constant human involvement. A prominent example of this approach is Reinforcement Learning from Human Feedback (RLHF)~\cite{DBLP:conf/nips/Ouyang0JAWMZASR22}, as illustrated in Figure~\hyperlink{prehoc_methods}{\ref{fig:prehoc_methods}(b)}. It first asks human annotators to label the preference for different LLM outputs and then train the reward model to predict the human preference. Afterward, reinforcement learning (RL) algorithms (\textit{e.g.}, Proximal Policy Optimization (PPO)~\cite{DBLP:journals/corr/SchulmanWDRK17}) are employed to optimize the model. RLHF and its variants have proven effective in correcting LLMs to become more beneficial and less harmful~\cite{bai2022training}, as well as instilling moral correctness~\cite{ganguli2023capacity}. 

% Plenty of works have been proposed to xxx
% however, human feedback is not the focus of xxx
% we briefly introduce the RLHF as a complement

\subsection{Learning with Automated Feedback}
% Collecting human feedback is generally resource-intensive, whether it is for optimizing LLMs or training the reward model. To minimize the demand for human intervention, numerous studies have explored the use of automated feedback for LLM correction during training time. 
Since collecting human feedback is quite resource-intensive, numerous studies have explored the use of automated feedback to minimize the demand for human intervention. 
% Define differences between reward model and automatic feedback
To differentiate between human and automated feedback, we define human feedback as a quality assessment performed by human evaluators on the outputs generated by the base model. This feedback is then used for either direct optimization or reward model learning (Section~\ref{subsec:learn_from_human_feedback}). On the other hand, automated feedback is collected in an offline environment, without the need for human assessment of model outputs. We mainly discuss training time strategies utilizing two types of automated feedback: \textit{extrinsic feedback} from external metrics/models, and \textit{intrinsic feedback} from the language model itself. 

\paragraph{External Metric Guidance.} Feedback provided by external metrics has been frequently used for training-time correction. Due to the discrete nature of metric signals, most approaches focus on non-differentiable training techniques. Minimum risk training~\cite{shen-etal-2016-minimum} optimizes model parameters with external evaluation metrics~\cite{xu-etal-2022-errors, xu-etal-2023-sescore2}, by incorporating metric score with maximum log-likelihood in the loss function. It can optimize metric scores during training time. However, it can lead the model to the robustness deficiencies of some metrics \cite{yan-etal-2023-bleurt}, such as BLEURT \cite{sellam-etal-2020-bleurt}. \citet{liu-liu-2021-simcls} leverages a contrastive learning framework to rerank candidates based on metric scores, which bridges the gap between training and inference objectives. \citet{li-etal-2019-deep} employs deep RL algorithm and \citet{unanue2021berttune} leverage Gumbel softmax \cite{jang2017categorical} to build distributional semantic reward from BERTScore \cite{bert-score} and mitigate exposure bias. To stabilize gradients, \citet{wu2021textgail} utilizes contrastive discriminator and PPO to imitate human texts. Recently, \citet{chang2023learning} propose a more efficient RL algorithm, RLGF, than PPO \cite{DBLP:journals/corr/SchulmanWDRK17} to finetune LLM with pre-defined reward. They integrate a reasonable but incomplete guide policy into a policy gradient framework and learn a near-optimal strategy. Different from leveraging feedback solely at fine-tuning, \citet{korbak2023pretraining} employs conditional training \cite{Keskar2019CTRLAC} and an automated classifier to tag undesirable contents at the pretraining stage. 

\paragraph{Self-Training.} Instead of leveraging external metrics as feedback, the language model itself can be used to provide feedback for its own output. This gives rise to the \textit{self-training} strategy of self-improving LLM by bootstrapping its original outputs, as depicted in Figure~\hyperlink{prehoc_methods}{\ref{fig:prehoc_methods}(c)}. \textit{STaR}~\cite{zelikman2022star} leverages the idea of CoT by prompting LLM to generate answers with rationales. By selecting rationales leading to the correct answer to further finetune LLM, the performance of LLM is improved. This process can be iterated with further performance gains. \citet{huang2022large} follows this idea by applying self-consistency \cite{wang2023selfconsistency} to majority vote reasoning paths (the paths that lead to the most voted answers). LLM is finetuned over selected reasoning-answer data with augmented prompts. This strategy has also been used to reduce the harmful responses of LLMs. \textit{RLAIF}~\cite{bai2022constitutional} adopted the strategy of critique $\rightarrow$ revision $\rightarrow$ supervised learning. The initial toxic responses are criticized and revised by the LLM itself following a set of human-defined principles. Afterward, the LLM is fine-tuned on the revised responses. \textit{AlpacaFarm}~\cite{dubois2023alpacafarm} further shows that LLMs can self-improve with RL. It designs LLM prompts to simulate human feedback in RLHF and shows that the feedback is effective and greatly reduces the cost. \citet{gulcehre2023reinforced} further improves self-training by proposing the \textit{Reinforced Self-Training} (ReST). It iteratively performs the following two steps to improve the LLM: 1) the \texttt{Grow} step produces a dataset by sampling from the policy model (\textit{i.e.}, the current LLM), and 2) the \texttt{Improve} step optimizes the LLM policy using offline RL algorithms. 

% \citet{scheurer2023training} uses LLM to generate feedback on its own output and select the best refinement from LLM that best incorporate output and feedback. Lastly, they finetune LLM by maximizing the likelihood of the chosen refinement. 

% Reference: Iterative-Optimization. Reasoning with Language Model Prompting: A Survey
% Self training -> Bootstrapping & Self-Improving: Towards Reasoning in Large Language Models: A Survey

% another line of work: train feedback model instead of LLM: Reflexion: Language Agents with Verbal Reinforcement Learning, The Unreliability of Explanations in Few-shot Prompting for Textual Reasoning, 

% \section{Feedback-Guided Generation}
\section{Generation-Time Correction}
\label{sec:feedback_guided_methods}

Correcting LLMs at training time appears to be the ideal solution given the principle of ``an ounce of prevention is worth a pound of cure''. However, there is no guarantee that all undesired behavior can be solved at training time. Moreover, training-time correction might be excessively resource-intensive or even impractical for many LLMs, \textit{e.g.}, closed-source LLMs where weights are inaccessible, and colossal LLMs with billions of parameters. This motivates the exploration of methods that seek to correct LLMs \textit{during} the generation time or \textit{after} the output is generated. This section focuses on generation-time correction techniques, in which automated feedback serves as a guiding mechanism for LLM generation. Such a strategy allows LLMs to rectify errors during generation without modifying the model parameters. We identify two primary strategies for generation-time correction: \textit{Generate-then-Rank}, and \textit{Feedback-Guided Decoding}. 

% In this section, we focus on the generation-time correction methods, where automated feedback is utilized as guidance for LLM generation, either for ranking the outputs or guiding the model to search for the optimal solution.

% This could be implemented for ranking potential outputs or guiding the model toward the discovery of optimal solutions. 

% This type of feedback, derived from interactions between LLMs and users in practical scenarios, enables models to learn from their errors and offers opportunities for ongoing refinement without altering model parameters. In addition, the feedback functions as a guiding mechanism, allowing the model to generate more desirable outputs by leveraging its existing capabilities. 

\begin{figure}[!t]
    \centering
    \hypertarget{in_generation_methods}{}
        \includegraphics[width=7.5cm]{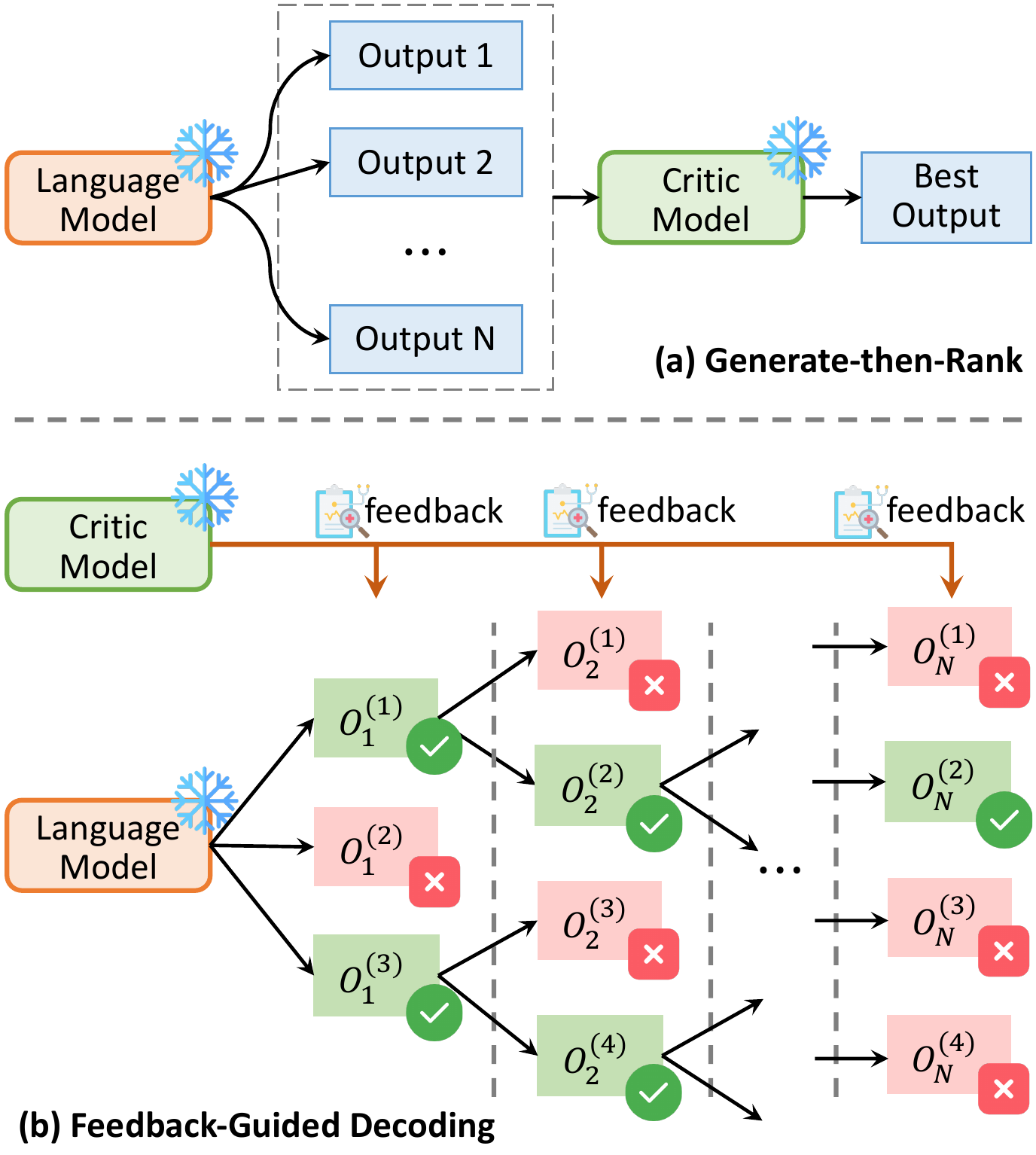}
    \caption{The illustrations of the two typical strategies of \textit{generation-time correction}: (a) Generate-then-Rank, and (b) Feedback-Guided Decoding.}
    \label{fig:in_generation_methods}
\end{figure}

% Generate-then-aggregate might be more accurate; or answer / output aggregation
\subsection{Generate-then-Rank}

The most immediate strategy involves sampling a large number of candidate generations and subsequently picking up the best generation based on the feedback provided by the critic model, as illustrated in Figure~\hyperlink{in_generation_methods}{\ref{fig:in_generation_methods}(a)}. Here, the critic model $\mathcal{C}$ aims to learn the mapping $x, \hat y_1, \cdots, \hat y_N \rightarrow y_{best}$, where $y_{best}$ is the best output among the $N$ candidate outputs $\hat y_1, \cdots, \hat y_N \sim \mathcal{M}(x)$.
% One effective method involves training reward models to discriminate between desirable and undesirable outputs.

This approach is often integrated with the Chain-of-Thought (CoT) prompting method~\cite{wei2023chainofthought} to tackle complex reasoning tasks, such as solving math word problems as in GSM8K~\cite{GSM8K}. Given an input problem $x$, the LLM initially generates multiple candidate solutions ${y_1, \cdots, y_n}$. Each solution $y_i = [z_i, a_i]$ comprises a reasoning path (explanation) $z_i$ leading to the predicted answer $a_i$. Subsequently, the critic model $\mathcal{C}$ assigns a plausibility score $s_i$ to each candidate reasoning path $z_i$. The final selection of the best solution from the scored set ${(z_i, a_i, s_i)}_{i=1}^n$ is achieved via either ranking or voting. 

Different critic models have been proposed in various works. For instance, DIVERSE~\cite{li-etal-2023-making} trains a binary verifier based on DeBERTa~\cite{DBLP:conf/iclr/HeLGC21}, using reasoning paths that correspond to the correct final answer as positive examples and the others as negative examples. The best answer is then determined by a majority vote of positively-verified candidates. \citet{weng2023large} introduced a training-free critic model based on the idea of self-verification, in which the plausibility score is calculated by assessing the consistency between the results of forward reasoning and backward reasoning. In a different vein, the RR~\cite{he2022rethinking} presented a critic model to assess the \textit{faithfulness} of each reasoning path by retrieving supporting information from a knowledge base. LEVER~\cite{ni2023lever} employed this strategy in language-to-code generation, with each solution $y_i$ serving as a candidate SQL program for the question $x$. A verifier was trained to predict the likelihood of a program's correctness based on the program itself and its execution results. A similar idea is adopted in CodeT~\cite{DBLP:conf/iclr/ChenZNZLLC23} where multiple code solutions and the test cases are generated by the LLM and the best code solution is selected by a dual execution agreement. 

% This strategy has been widely adopted in the arithmetic reasoning task, represented by solving the math word problems in GSM8K~\cite{GSM8K}. 
% The DIVERSE~\cite{li-etal-2023-making} model trained a critic model to assign a plausibility score to each candidate reasoning path. 

\subsection{Feedback-Guided Decoding}
% Verifier-guided search (external verifier is guided by the LLM reasoning/CoT and can guide self-repair)

The generate-then-rank method, in which the critic model offers \textit{output-level feedback} on the entire reasoning path, has certain limitations: 1) The output-level feedback is not fine-grained enough to pinpoint the exact error locations, 2) The extensive length of the output can complicate its quality assessment, and 3) This method does not facilitate fine-grained control over the generation process. For example, the LLM cannot correct its errors during the generation process but must wait until the entire output has been generated. 

To address these issues, several works have adopted the \textit{feedback-guided decoding} strategy shown in Figure~\hyperlink{in_generation_methods}{\ref{fig:in_generation_methods}(b)}, which relies on \textit{step-level feedback} to offer fine-grained guidance over the generation process. Here, the generation of the output $y$ is broken down into multiple reasoning steps (or thoughts), \textit{i.e.}, $y_i = [o_1, o_2, \cdots, o_n]$. % At each individual reasoning step $i$, the critic model provides feedback on the intermediate outputs $[o^{(1)}, \cdots, o^{(i)}]$ produced so far. 
At each individual reasoning step $t$, the critic model provides feedback $\mathcal{C}(x, o_{1:t-1}, o_t)$ that indicates the quality of $o_t$ as a candidate step. With the ability to generate and evaluate individual steps, a search algorithm, such as beam search or depth-first search, can be employed for a systematic exploration of the output space, which effectively steers the decoding process toward the generation of an optimal solution. This also allows the LLM to recover from its early mistakes during generation and helps alleviate the \textit{reasoning inconsistency} problem~\cite{zelikman2022star,DBLP:SelectionInference}, \textit{i.e.}, incorrect reasoning leads to correct final answer. 

The feedback-guided decoding strategy has been applied in many recent works, such as \textit{Tree-of-Thought}~\cite{yao2023tree}, \textit{GRACE}~\cite{KhalifaGRACE}, and \textit{RAP}~\cite{hao2023reasoning}. 
% \textit{SelfEval-Guided-Decoding}~\cite{xie2023decomposition}. 
These works mostly differ in how to obtain the critic model that provides automated step-level feedback, the most challenging but crucial element of this strategy. We classify their employed methods into four categories: human feedback, a trained verifier, external metrics, external knowledge, and self-evaluation. 

\vspace{0.15cm}

% \noindent $\bullet$ \textbf{Reward Model from Human Feedback.} One way is to train a step-level reward model by collecting human feedback, similar to the methods introduced in Section~\ref{subsec:learn_from_human_feedback}. \citet{uesato2022solving} ask human annotators to evaluate the correctness of each reasoning step for the problems in GSM8K and then train a binary reward model. \citet{lightman2023lets} further annotated a larger 800K dataset of human step-level feedback. Both papers found that step-level feedback helps to train a more reliable reward model and can largely improve the faithfulness of the generated reasoning path. 

\noindent $\bullet$ \textbf{Reward Model from Human Feedback.} One approach involves training a step-level reward model by gathering human feedback, much like the methods discussed in Section~\ref{subsec:learn_from_human_feedback}. \citet{uesato2022solving} ask human annotators to evaluate the correctness of each reasoning step for the problems in GSM8K and subsequently train a binary reward model. \citet{lightman2023lets} expand this approach by annotating a larger dataset consisting of 800K instances of human step-level feedback. Both studies discovered that step-level feedback assists in training a more reliable reward model, which enhances the faithfulness of reasoning. 

\vspace{0.15cm}

\noindent $\bullet$ \textbf{Training Verifier with Synthetic Data.} Considering the high cost of collecting human annotations and their limited scalability, some works~\cite{DBLP:conf/emnlp/Yang0C22,DBLP:conf/emnlp/TafjordMC22,li-etal-2023-making,KhalifaGRACE} have trained a step-wise verifier using automatically constructed training data. Positive examples are derived from ground-truth reasoning paths, while negative examples are synthesized by proposing an alignment algorithm~\cite{KhalifaGRACE} or by making text perturbations on positive samples~\cite{DBLP:conf/emnlp/Yang0C22}. 

\vspace{0.15cm}

\noindent $\bullet$ \textbf{Feedback from External Metric.} Several works also leverage external metrics to re-rank or guide text generation. \citet{freitag-etal-2022-high} uses minimum bayes risk decoding on unbiased samples to optimize neural metrics as an alternative to beam search. Plug and play~\cite{Dathathri2020Plug} combines a pretrained model with attribute classifiers that guide text generation without any further training of the model. It leverages the gradient of the classifier to update LM and increase the likelihood of the desirable attribution at the text generation of LM. FUDGE~\cite{yang-klein-2021-fudge} reweights the model predictions at each token and estimates the attribution classification at each partial sequence. Following up to the gradient-based approach, DiffusionLM \cite{Li-2022-DiffusionLM} obtains a sequence of intermediate latent variables by denoising a sequence of Gaussian vectors. It performs the iterative gradient updates over latent representations to satisfy controlled requirements from an attribute classifier. 

\vspace{0.15cm}

\noindent $\bullet$ \textbf{Feedback from External Knowledge.} External knowledge sources have also been used to guide the LLM in generation. \citet{varshney2023stitch} retrieves relevant knowledge from Wikipedia as evidence to validate and correct LLM's generated sentences at each step. Once a non-factual sentence is corrected, the revised sentence is added back to the input along with the prior generations to continue generating the next sentence. In a different approach, MemPrompt~\cite{madaan-etal-2022-memory} leverages prior user feedback as a knowledge source. It maintains an external pool of user feedback and searches it for responses that match the intent of the current query. The retrieved feedback is then concatenated with the input to guide the following generation. 

% External knowledge and memory ???
% ed as a source of feedback to detect and revise factual errors in LLM’s output and to support LLM-generated facts with evidence or citations. 

\vspace{0.15cm}

% \noindent $\bullet$ \textbf{Training Verifier with Synthetic Data.} Since human annotations are expensive to collect and do not scale well, some works~\cite{DBLP:conf/emnlp/Yang0C22,DBLP:conf/emnlp/TafjordMC22,li-etal-2023-making,KhalifaGRACE} trained the step-wise verifier by automatically constructing training data. Positive examples can be extracted from ground-truth reasoning paths. Negative examples are automatically constructed via alignment algorithm~\cite{KhalifaGRACE} or text perturbation~\cite{DBLP:conf/emnlp/Yang0C22}. 

% \noindent $\bullet$ \textbf{Self-Evaluation.} Some works adopted a more flexible way that using the LLM itself as the critic model by giving appropriate prompts. In Tree-of-Thought~\cite{yao2023tree}, the LLM is prompted to evaluate the value of the current state by outputting a scalar value (\textit{e.g.} 1-10) or short words (\textit{e.g.}, sure/likely/impossible). \citet{xie2023decomposition} adopted a similar strategy to prompt the LLM with ``Is the above step of reasoning: (A) Correct (B) Incorrect''. 

\noindent $\bullet$ \textbf{Self-Evaluation.} Some studies have utilized a more flexible strategy, employing the LLM itself as the critic model by designing appropriate prompts. For instance, in Tree-of-Thought~\cite{yao2023tree}, the LLM is prompted to assess the value of the current state by producing a scalar value (\textit{e.g.}, 1-10) or short phrases (\textit{e.g.}, sure/likely/impossible). \citet{xie2023decomposition} employed a similar approach by prompting the LLM with ``Is the above step of reasoning: (A) Correct (B) Incorrect''. Self-evaluation provides an efficient evaluation method without requiring task-specific verifier fine-tuning. 

\vspace{0.15cm}

Existing works also adopted different strategies to control the decoding process with the help of the step-level critic model. Tree-of-Thought employed breadth-first search and depth-first search, while GRACE~\cite{KhalifaGRACE} and \citet{xie2023decomposition} adopted the beam search strategy. At each step, the top-$k$ scoring candidates are selected for subsequent generations. This process is repeated until the final answer is generated. Instead, CoRe~\cite{DBLP:conf/acl/ZhuWZZ0GZY23} and RAP~\cite{hao2023reasoning} adopted the Monte Carlo Tree Search (MCTS) to strike a proper balance between exploration and exploitation to find the best reasoning path more efficiently.

\begin{figure*}[!t]
    \centering
    \hypertarget{posthoc_methods}{}
        \includegraphics[width=16cm]{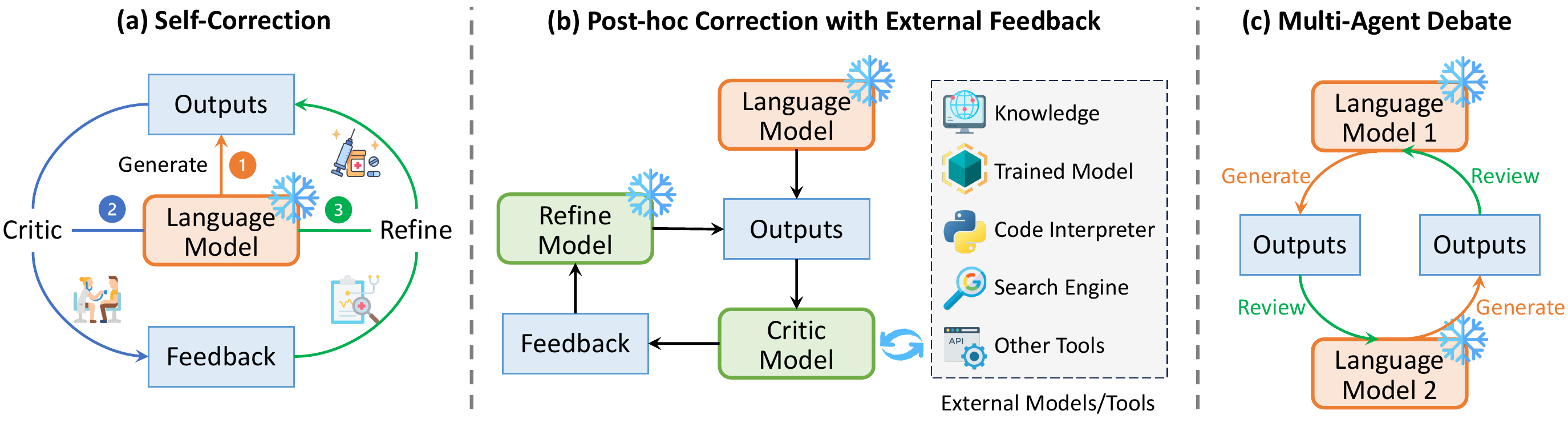}
    \caption{Three typical strategies of \textit{post-hoc correction}: self-correction (a), post-hoc correction with external feedback (b), and multi-agent debate (c).}
    \label{fig:posthoc_methods}
\end{figure*}

\section{Post-hoc Correction}
\label{sec:posthoc_methods}
% While the generation-time correction strategy offers great flexibility and empowers the model to identify and rectify its errors early on, its success heavily depends on the critic model providing accurate feedback for intermediate outputs. 

The success of generation-time correction heavily depends on the critic model's capability in providing accurate quantifiable feedback for intermediate outputs. However, this might be quite challenging for many NLP tasks, such as summarization, due to the holistic nature of the evaluation, \textit{i.e.}, the summary can only be accurately assessed after the entire summary is generated. This motivates the employment of \textit{post-hoc correction} methods, where both the critic and refine models intervene only \textit{after} the entire output is produced. Post-hoc correction also provides a more effective interface with various forms of insightful natural language feedback. This feedback can be as detailed as a diagnostic report pinpointing exact error locations, or as general as suggestions for overall writing improvement. As illustrated in Figure~\ref{fig:posthoc_methods}, we summarize three primary strategies for post-hoc correction: \textit{Self-Correction}, \textit{Correction with External Feedback}, and \textit{Multi-Agent Debate}. 

\subsection{Self-Correction}
The simplest approach to implement post-hoc correction is the ``Self-Correction'' technique, where an LLM is employed to generate feedback and refine its own output. As depicted in Figure~\hyperlink{posthoc_methods}{\ref{fig:posthoc_methods}(a)}, an LLM is initially used to produce a initial output, and subsequently, the same model acts as a critic to generate feedback and refine this initial output based on the received feedback. This process is typically iterative and continues until an output of acceptable quality is obtained or a pre-specified number of iterations are reached. 

% PEER\citep{schick2022peer} uses a slightly modified version of above approach to create a collaborative text-editing model. PEER proposes to train a model which imitates a text writing process: planning, suggesting changes and editing the draft. Essentially given an initial text and supporting documents, authors train an autoregressive model to first generate a high-level plan expressed as a short sentence ("fix grammatical errors", "add more information" etc.), then use a given set of documents for editing the text according to the plan and optionally generate an explanation for the edit. 
% While this approach seems intuitive, it suffers from data scarcity as it is hard to capture this editing process using natural means of data collection, thus making it hard to train these models accurately. 

% \dn{I don't think PEER counts as self-correction, maybe find a better section for this work?}

\textit{Self-Refine}~\citep{madaan2023selfrefine} proposed a simple-yet-effective self-correction framework by simply using a single powerful pre-trained LLM to generate output, provide feedback, and refine the output based on that feedback. 
% All these steps are achieved by re-purposing the same LLM with different prompts. 
All these steps are conducted using the same LLM, guided by different prompts.
Similarly, \textit{Clinical Self-Verification}~\citep{gero2023selfverification} employs the self-correction framework to extract patient data from clinical notes. They specifically generate feedback to find missing elements in the initially extracted data and to validate the generated data. The output is then refined by eliminating unsupported elements. In contrast, \textit{Reflexion}~\citep{shinn2023reflexion} highlighted that prior self-correction research focused on single-turn generation tasks and failed to retain a record of past errors. To address this, \textit{Reflexion} proposed to use the same self-correction framework with an addition of a ``long-term memory'' capable of storing prior feedback and outputs, thereby avoiding the repetition of previous mistakes. Also, \textit{Reflexion} improves \textit{Self-Refine} by incorporating scalar-valued feedback and other forms of feedback. 

While self-correction has shown effective for a wide variety of text-generation tasks, this strategy requires the use of powerful, large-scale LLMs capable of refining text based on provided feedback. As noted by \citet{madaan2023selfrefine}, smaller, open-source models often struggle to refine their output effectively, even when the correct feedback is provided. A possible solution involves explicitly training models for this self-correction process. \textit{SelFee}~\citep{selfee2023} proposes training a model to emulate the self-correction process by generating output, feedback, and a refined solution in an auto-regressive manner. They use more powerful LLMs to provide feedback and refinement data, with data collection facilitated through ChatGPT. 

\subsection{Models/Tools as Feedback}
% Tool-assisted Correction
As self-correction relies on the language model for feedback, the quality of the feedback is inherently constrained by the inherent limitations of LLMs, such as the inability to access up-to-date information, take actions, or perform precise mathematical reasoning. To address this, recent works have investigated the use of external tools for providing feedback. Illustrated in Figure~\hyperlink{posthoc_methods}{\ref{fig:posthoc_methods}(b)}, a broad array of external tools, including trained models, code interpreters, and search engines, can be integrated to provide specialized feedback. 

\paragraph{Code Interpreter.} In code generation, the program executor is frequently used as a source of feedback for refining the initial code written by the model. For example, \textit{Self-Edit}~\cite {zhang2023selfedit} and \textit{Self-Evolve} execute the initial program on example test cases and provide the execution results back as feedback. Afterward, an LLM is prompted to refine the initial code based on the feedback. \textit{Self-Debug}~\cite{chen2023teaching} investigated using program explanation, unit tests, and program interpreter as feedback types. \textit{ALGO}~\cite{zhang2023algo} explored a more fine-grained feedback for code generation. For each problem, it first generates a reference oracle program that solves the problem with an exhaustive search. The feedback is collected by comparing the outputs from the LLM-generated program with the oracle outputs for a given set of test inputs. The self-correction strategy has also been adopted for formal verification of software. \citet{charalambous2023new} employed Bounded Model Checking to locate the software vulnerability and then used the LLM for correction. 

\paragraph{Logic Reasoner.} Tool-assisted feedback has also been used to enhance the faithfulness of LLMs' reasoning. For example, Logic-LM~\cite{pan2023logiclm} solves a logical reasoning problem by first translating it into logical form with LLMs and then performing inference on it with external symbolic solvers. Due to the complexity of correctly parsing the problem at the first attempt, a self-refinement module is introduced to modify inaccurate logical forms using the error messages returned by the symbolic reasoner as feedback. Similarly, Baldur~\cite{first2023baldur} uses existing search-based proof assistants as a source of feedback to improve language models' ability to generate theorem proofs. 

% RARR
% LLM-Augmenter
% Self-Checker
% REFEED
% FacTool
\paragraph{External Knowledge.} External knowledge is also frequently incorporated as a source of feedback to detect and revise factual errors in LLM's output and to support LLM-generated facts with evidence or citations. For example, \textit{RARR}~\cite{gao2023rarr} and \textit{REFEED}~\cite{yu2023improving} directly prompt LLMs to raise questions about different aspects of the generated output. An external retriever then searches for evidence to investigate each query. Finally, a refine model is employed to amend the output based on any detected discrepancies between the output and the retrieved evidence. \textit{LLM-Augmenter}~\cite{peng2023check} proposes a similar method but differentiates itself by automatically generating natural language feedback based on the evidence retrieved. This feedback identifies error locations and provides revision suggestions. These models are evaluated on knowledge-intensive QA tasks. To broaden the adaptability, \textit{FACTOOL}~\cite{chern2023factool} extends knowledge-assisted factual error correction to a wider range of tasks, including code generation, mathematical reasoning, and scientific literature review. 

\paragraph{Trained Model.}
Existing works also fine-tune specialized models for feedback generation. These critic models can then be paired with similar or more potent language models in an iterative-refinement cycle. For example, \textit{CodeRL}~\cite{le2022coderl} treats program synthesis as a reinforcement learning task and trains a critic model whose output optimizes the main model. In contrast, \textit{REFINER}~\cite{paul2023refiner} trains a task model to produce an intermediate representation for problem-solving, while a critique model provides feedback on each intermediate training step. The critique model can subsequently be employed to generate feedback for larger task models, such as ChatGPT. Similarly, \textit{RL4F}~\citep{DBLP:conf/acl/AkyurekAKCWT23} utilizes reinforcement learning for training a critic while keeping the downstream task model fixed. The critic model is initially fine-tuned to produce feedback given an initial output and then further fine-tuned using a policy optimization method. The reward is defined by evaluating the accuracy of the refined output or comparing it with the ground truth, assuming that the downstream model can effectively refine the output if accurate feedback is provided. In ``red-teaming'' type applications, including targeting vulnerabilities in the content filtering capabilities of other generated systems, feedback from the content filters themselves can be used as a signal to guide the generation of better adversarial examples. For example, Feedback Loop In-context Red Teaming (FLIRT) \cite{mehrabi2023flirt} uses the signal from an explicit image classifier to guide an LLM to produce adversarial input prompts for a text-to-image system to generate more unsafe images for auditing purposes. 

\paragraph{Integrating Multiple Tools.}
Broadening the idea of tool-assisted feedback, \textit{CRITIC}~\citep{Gou2023CRITICLL} integrates a variety of tools in a unified framework, including program interpreters for coding feedback, external knowledge and search engines for factual information, calculators for verifying mathematical equations, and LLM-based natural language feedback. Each tool is proficient at providing feedback for different aspects, contributing to a more comprehensive feedback system. 

\subsection{Multi-Agent Debate}
Besides integrating external tools, recent studies have also explored the strategy of \textit{debating between multiple LLMs}, drawing inspiration from collaborative intelligence, where diverse perspectives often converge to a more refined solution. This approach aims to improve the output quality by employing multiple instances of LLMs, each proposing and debating their individual responses over multiple rounds to arrive at a common final answer. 

\citet{du2023improving} first applied and evaluated this strategy in arithmetic reasoning tasks. Each agent (a duplicate of LLM) initially generates its individual solution along with justifications. The debate phase involves collating responses from all agents and presenting this as context to each agent. Based on this context, each agent is then instructed to craft a revised response. The models are found to converge on a shared solution following multiple debate iterations. Experiments show that multi-agent debate leads to improved performance over the self-correction strategy. Furthering this concept, \textit{PRD}~\cite{li2023prd} proposed the peer rank algorithm for better obtaining a consensus answer after debating. It considers pairwise preferences between all possible answer pairs from individual LLMs and uses these preferences to generate a final ranking of models. 

In addition to reasoning tasks, \textit{LM vs LM}~\cite{cohen2023lm} further demonstrated the effectiveness of multi-agent debate for detecting factual errors. The approach involves a generator LLM creating a claim, while an examiner LLM probes for factual inaccuracies through a multi-turn interaction.To broaden this concept, \citet{fu2023improving} demonstrated that interactions between different LLMs could mimic human behavior in real-world tasks. The study showcased this through a bargaining scenario where different LLM agents assumed the roles of buyer and seller. This further highlights the versatile applications of multi-agent debates. 

% The strategy of relying on a single LLM for self-reflection is often constrained by the inherent capabilities of the model itself. Drawing inspiration from collaborative intelligence, where diverse perspectives converge to solve complex problems, researchers have begun to investigate the strategy of \textit{debating between multiple LLMs}. 

% and deliberating their individual responses and reasoning processes. This iterative exchange occurs over multiple rounds, leading to a mutually agreed-upon final answer. 

\section{Applications}
\label{sec:applications}

%Previously we outlined a variety of proposed automated correction mechanisms for large language models, grouped by implementation commonalities.
%Here we instead group by end-use application area, to elucidate.

Following our above outline of automated correction techniques, we now will briefly discuss the various application domains for which automated correction is useful, and point out commonalities in self-correction strategies, and discuss how improvements to performance in self- or feedback-driven correction will give rise to downstream performance improvements.

\subsection{Factual Correction}

Many of the aforementioned automated correction strategies implicitly engage factual correction to improve various errors in output \cite{gao2023rarr,peng2023check}.
It is therefore natural that these capabilities can be directly engaged for factuality detection and factual correction as an end task, such as in \textit{LM vs LM} \cite{cohen2023lm} or \textit{Multiagent Debate} \cite{du2023improving}.
When external tool use is acceptable, retrieved facts can be leveraged to further improve the factuality of generated text in a self-correcting manner without pure reliance on memorized knowledge \cite{Gou2023CRITICLL}.

In short, self-correcting is a foundational technique in many LLM-based fact correction or fact-checking systems, and \textit{models that are better at self-directed factual correction will probably be better in a host of other self-correction settings}.

% The improved version of this and next subsection should cite commonsense literature

\subsection{Reasoning Tasks}

In most \textit{reasoning tasks}, no good references from which outputs can be sanity-checked are readily available \cite{choi2023commonsense}. This is unfortunate, as the ``reasoning'' capabilities provided by LLMs---in particular, their ability to \textit{operate on} natural language from \textit{instructions specified in} natural language---is a core driver for their popularity. \textit{Reasoning tasks} constitute a broad array of problems where this capacity is most necessary. For example, more complex multi-hop question answering tasks \cite{yang2018hotpotqa,chen2021finqa} can be construed as requiring both factual correction and reasoning correction capabilities \cite{ho2023wikiwhy}. Thus the question becomes, \textit{how can we prompt the LLM to identify and correct intermediate reasoning errors?} Similarly to other application areas, both pure-LLM-driven and external tool-based error detection techniques have been proposed. 
% inconsistency detection and correction

LLM-based implementations of reasoning error detection include debate-based techniques, which can be thought of as implicitly rolling consistent reasoning enforcement in the ``critic'' module \cite{cohen2023lm,li2023prd}, and self-refinement techniques \cite{manakul2023selfcheckgpt}.
In short, a given passage exhibiting reasoning (eg, step-by-step \cite{wei2023chainofthought}) is fed into an LLM, which is prompted to check for and/or correct reasoning errors directly. The error detection often collaborates with a decoding algorithm such as the beam search to lead the reasoning towards the correct direction~\cite{hao2023reasoning, yao2023tree}.  

External feedback using techniques such as natural language inference (NLI) can be both directly leveraged to spot errors as a heuristic for correction, and as a means to score the quality \cite{DBLP:conf/emnlp/Yang0C22,DBLP:ROSCOE}. 
However, there are some open questions regarding the quality of the supervised learning-based tools like NLI \cite{srikanth2022partial,saxon2023peco}. 

Among different types of reasoning, the self-correction strategy has been well studied and implemented for \textit{arithmetic reasoning}, as outlined in Table~\ref{tbl:summary_works} and Table~\ref{tbl:summary_works_2}. One of the reasons for this skew is the relative ease of verifying intermediate reasoning steps within arithmetic problems. Some recent studies~\cite{pan2023logiclm,first2023baldur} have started to extend the application of this strategy to deductive reasoning. However, the implementation of self-correction in a wider array of reasoning tasks, including inductive and abductive reasoning, is still relatively under-explored. 
% , indicating promising directions for future research. 

\subsection{Code Synthesis}

Code generation is a burgeoning application domain for LLMs for which correction is particularly important. Even human programmers tend to write code through an iterative process of addition and correction. For humans, strategies such as reading linter warnings, compiler/runtime errors, and incorrect outputs to diagnose necessary changes to the source are all employed in the software development process. Each of these strategies has natural analogues in the code generation pipeline.

The aforementioned warnings, errors, or outputs are usually fed directly back into the LLM to guide the code correction process~\cite{zhang2023selfedit,DBLP:conf/iclr/ChenZNZLLC23}. After all, compiler failures are a particularly strong signal that a piece of code will not work, having great utility in guiding LLM self-correction. More excitingly, other work proposed to utilize more fine-grained feedback such as program explanations generated by LLMs~\cite{chen2023teaching} and the comparison with a reference oracle program~\cite{zhang2023algo}.
% in code in natural language and the results are promising, such as in \ms{example \tocite}.

%  However, only very limited studies on how and when self-repair works effectively exist in the literature, and one might wonder to what extent a model is really capable of providing accurate feedback on why the code is wrong when that code was generated by the same model. 

% With this evaluation strategy, we find that the effectiveness of self-repair is only seen in GPT-4. We also observe that self-repair is bottlenecked by the feedback stage; using GPT-4 to give feedback on the programs generated by GPT-3.5 and using expert human programmers to give feedback on the programs generated by GPT-4, we unlock significant performance gains.

The above works only \textit{empirically} show that LLMs exhibited a remarkable capability for self-repairing the codes. In recent work, \citet{olausson2023demystifying} further conducted an in-depth analysis of how and when self-repair works effectively. They found that self-repair is bottlenecked by the feedback stage: substantial performance improvements were only noticed when feedback was provided by expert human programmers or GPT-4. This revelation raises intriguing questions, such as whether self-repair is an emergent ability exclusive to certain LLMs and how could we endow smaller models with similar capabilities. 

% However, hooking up the outputs of an LLM directly into an \texttt{eval()} function can introduce security risks \tocite, as many adversarial attack vectors exist to produce arbitrary undesired behavior from LLMs \tocite. Thus, it is extremely advisable to place strict limits around the sort of execution that is allowed for LLM-generated code in an automated pipeline.

\subsection{Other Applications}

% \paragraph{Creative Generation.} 
\paragraph{Open-ended Generation.} 
In addition to handling pure factuality issues, self-correction can be applied to subjective qualities of the generated text. These interventions include post-hoc self-correcting \textit{toxic / harmful outputs}~\cite{Gou2023CRITICLL,DBLP:conf/iclr/WelleckLWBSK023,helbling2023llm}, enhancing the narrative quality in \textit{story generation}~\cite{yang-etal-2022-re3}, and refining \textit{response generation} in dialogues~\cite{selfee2023,yu2023improving}. Given the subjectivity involved in assessing the outputs, these studies often rely on detailed, natural language feedback and employ an iterative refinement strategy for post-hoc refinement. 

\paragraph{Machine Translation.}
The concept of post-hoc self-correction has deep roots in the field of machine translation (MT), where it is often called \textit{Automatic Post-Editing} (APE)~\cite{DBLP:journals/mt/CarmoSMWHPSGW21}. A long line of prior works train models to fix translation errors by either learning from human correction data~\cite{DBLP:conf/eacl/AlabauBCCGGGHKLMOSST14} or from synthetic training data~\cite{lee-etal-2021-adaptation}. To minimize the cost of data collection, recent works~\cite{Chen2023IterativeTR,raunak2023leveraging} have leveraged the in-context learning ability of LLMs for post-editing translations. Besides post-hoc methods, training-time correction~\cite{unanue2021berttune} and decoding-time correction~\cite{freitag-etal-2022-high} are also adopted by prior works. 

%MBR Decoding~\cite{freitag-etal-2022-high}
%BERTTune~\cite{unanue2021berttune}
%IterRefinement~\cite{Chen2023IterativeTR}
%Auto-Post-Editing~\cite{raunak2023leveraging}

\paragraph{Summarization.}

Automated model correction is commonly used in summarization to ensure the \textit{factuality} of the generated summary. There are two mainstream methods: 1) training-time correction that imposes factuality constraints during training~\cite{liu-liu-2021-simcls,DBLP:conf/naacl/WanB22,scheurer2023training}, and 2) post-hoc correction that post-edits generated summaries to correct factual errors~\cite{cao-etal-2020-factual,Saunders2022SelfcritiquingMF}. Recent works have investigated using RL to refine the model guided by automated feedback from either reward models~\cite{DBLP:conf/acl/AkyurekAKCWT23} or language models~\cite{pang2023language}. 

% SimCLS~\cite{liu-liu-2021-simcls}
% ILF~\cite{scheurer2023training}
% SIRLC~\cite{pang2023language}
% RL4F~\cite{DBLP:conf/acl/AkyurekAKCWT23}
% Self-Critique~\cite{Saunders2022SelfcritiquingMF}

% \paragraph{Multimodal settings.} A frontier of LLM development is in creating multimodal models that can take both images and text as input \tocite. Many of the aforementioned language-only tasks such as question answering, reasoning, and creative conditional generation can be expanded to the language and vision domain to provide the same outputs with additional image conditioning. 

% This opens up novel avenues for

% \paragraph{Embodied agents.}
% \ms{In reference to ``text is the universal interface'' paper:} open-ended manipulation in the medium of natural language can enable almost any computing task. We expect many of the findings from prior work to generalize to novel problem domains given they can be expressed naturally in text. Some potential difficulties seem to arise when ``reasoning'' is hindered by the unnaturalness of the textual representation (eg, loss of structure when converting graphs into textual representations leading to poorer ability to reason on graphs or inability to deal with blockworld problems when the blockworld has to be converted into plaintext). Future multimodal systems might face far fewer limitations in this domain.

% \section{Current Understandings, Limitations, and Future Directions}
\section{Research Gaps and Future Directions}
\label{sec:discussion}

% Large language models (LLMs) have exhibited a remarkable capability for self-analysis and self-improvement, as highlighted by numerous studies [28, 20, 13].
% These models can generate feedback on their initial output and subsequently use this feedback to refine their results.
% However, what's the limitation? what's the upper bound? there lacks a theoretical analysis on this. 
% although there are some empirical studies, such as Demystifying

%  However, only very limited studies on how and when self-repair works effectively exist in the literature, and one might wonder to what extent a model is really capable of providing accurate feedback on why the code is wrong when that code was generated by the same model. 

% With this evaluation strategy, we find that the effectiveness of self-repair is only seen in GPT-4. We also observe that self-repair is bottlenecked by the feedback stage; using GPT-4 to give feedback on the programs generated by GPT-3.5 and using expert human programmers to give feedback on the programs generated by GPT-4, we unlock significant performance gains.

\subsection{Theoretical Justifications}
Although LLMs have exhibited a remarkable capability for self-analysis and self-improvement, there remains a lack of theoretical justifications to uncover the mystery of such ability. Therefore, we argue that the study of underlying theoretical principles can offer a more transparent understanding of self-correction. Subsequently, we propose several potential directions for such explorations. 

The ability of language models to self-correct is closely associated with their capacity to exhibit metacognitive awareness, \textit{i.e.}, their understanding of their own knowledge and uncertainties \citep{kadavath2022language}. Similarly, the notion of \textit{calibration} in language models, referring to their ability to produce well-calibrated predictions with probabilities aligning closely with observed frequencies of outcomes, is of paramount importance \citep{lin2023generating}. Recent research by \citet{kadavath2022language} reveals that pre-trained language models, when presented with properly formatted multiple-choice and true/false questions, demonstrate good calibration. Particularly, language models exhibit well-calibrated responses to self-evaluation questions in few-shot settings. On the other hand, fine-tuned language models, such as those incorporating RLHF, require temperature adjustments to achieve calibration since the model distribution is tailored to optimize specific behaviors. 

While language models demonstrate some capacity for self-feedback, achieving superior performance often necessitates incorporating external feedback signals. The integration of feedback signals is closely linked to the alignment of language models, a domain that still lacks comprehensive understanding. For example, in RLHF, the choice of the metric to minimize between the reward model output and the final model output significantly impacts downstream task performance \citep{go2023aligning}, yet this aspect remains underexplored in many applications. Furthermore, the optimal method for automatically generating prompts to instruct language models effectively, for tasks such as evaluating and refining their outputs, remains an open question. Although \citet{sordoni2023deep} have addressed this issue by treating natural language prompts as parameters of the language model and performing discrete optimization, their approach still requires hand-crafted meta prompts to implement the algorithm.

% \subsection{Measuring and Improving Self-Correction Ability}
\subsection{Measuring the Ability of Self-Correction}

Despite the promising progress in enabling LLMs to self-correct and improve their outputs, current research only provides \textit{empirical} evidence of its effectiveness across diverse applications. However, a gap remains in establishing robust quantitative metrics to understand and evaluate the self-correction capability of LLMs. While various strategies have been proposed to rectify LLM outputs, a comprehensive comparative evaluation of these approaches is still missing. This includes metrics to assess the effectiveness, applicability, complexity, and potential upper-bound limits of each strategy within a unified context. Future research could aim to create comprehensive evaluation frameworks which take into account variables such as the complexity of the task, the degree of initial error, the improvement in quality after self-correction, etc. 

In addition to this, building benchmarks to diagnose self-correction capabilities represents another promising research direction. Such diagnostic datasets would enable a more standardized, objective evaluation of different LLMs and self-correction strategies, driving the development of more accurate and efficient models. 

% In addition, there lack studies aiming to improve the self-correction ability of LLMs. Although different 
% there is still much room for future research to measure 

\subsection{Continual Self-Improvement}
% life-long learning
Another promising yet under-explored area of LLM self-correction is the idea of continual, life-long self-improvement. As LLMs are utilized in more diverse, dynamic, and real-time contexts, the ability to adapt and improve continually over time becomes essential. This is closely related to the concept of continual (life-long) learning ~\cite{DBLP:lifelong}, where the model continually learns new skills and adapts to novel environments and contexts. Translating this to self-correction implies that LLMs continuously evaluate their outputs, learn from errors, update their knowledge, and adapt their decision-making strategies accordingly. 

Studies on self-training such as ~\cite{huang2022large,zelikman2022star} have evidenced that LLMs can self-improve by continuously training on their own outputs that are positively evaluated by humans or models. However, these studies typically concentrate on a single, one-time correction process and evaluate improvements in a particular aspect. The robustness and stability of self-training under continual settings remain uncertain. For example, a major challenge of continual learning is catastrophic forgetting~\cite{DBLP:journals/corr/KirkpatrickPRVD16}, where the acquisition of new skills often leads to a considerable decrease in previous capabilities. It is unclear whether similar issues may emerge in a continual self-improve LLM, such as whether correcting one behavior may unintentionally alter a previously corrected behavior. 

Finally, exploring how to integrate various self-correction techniques to efficiently build a continual self-improvement LLM is also worth investigating. For example, post-hoc correction represents a more immediate and less costly strategy, while training-time correction addresses model behavior more fundamentally, with a higher computational cost. To combine these strategies, post-hoc correction could be used to collect training data (\textit{e.g.}, most frequently made mistakes and their corrections), which are used to guide the periodically training-time correction to address these recurring issues permanently. 
% This hybrid approach could enable continual self-improvement while balancing effectiveness and efficiency.

\subsection{Self-Correction with Model Editing}

Recent years have witnessed a surge in techniques for \textit{model editing}~\cite{DBLP:conf/iclr/SinitsinPPPB20,DBLP:conf/emnlp/CaoAT21,DBLP:model_edit}, aiming to adjust the model's behavior for examples within the editing scope while leaving its performance for out-of-scope examples unaltered. Model editing has been employed in updating outdated knowledge embedded in LLMs~\cite{DBLP:conf/acl/0002HHLPL22,DBLP:conf/acl/OnoeZPDC23} and in addressing issues related to false associations memorized during LLM training~\cite{DBLP:conf/emnlp/MurtyMLR22,DBLP:conf/nips/TannoPNL22}. Current model editing methods have shown some efficacy in adjusting factual knowledge within LLMs, yet they still suffer problems such as a lack of robust generalization capabilities~\cite{DBLP:model_edit} and the introduction of substantial unintended side effects~\cite{DBLP:conf/acl/Hoelscher-Obermaier23}. 

Nevertheless, the advancements in model editing present promising opportunities for the self-correction of LLMs. Primarily, model editing enables accurate, fine-grained corrections at the level of individual neurons or layers, circumventing the need for extensive retraining associated with training-time correction. Moreover, through the analysis of the impact of model edits, we can deepen our understanding of the self-correction mechanism. Further, methods developed to curtail undesired side effects in model editing~\cite{DBLP:conf/acl/Hoelscher-Obermaier23} could foster more robust self-correction strategies by mitigating the issue of inadvertently introducing new errors while resolving existing ones. Therefore, we forecast future research to incorporate model editing into LLM self-correction processes, an under-explored area. 

% \subsection{Combining Automated Feedback with Human Feedback}
% \subsection{Human-in-the-loop Automated Feedback}
% \subsection{Automated + Human Feedback}
% human feedback on automated feedback 
% learn feedback model?

% \subsection{Multimodal Extensions}
% \subsection{Multimodal Feedback Learning}
% \subsection{More Tasks and Settings}
\subsection{Multi-Modal Self-Correction}

As discussed in Section~\ref{sec:applications}, self-correction strategies for LLMs have been successfully employed in an extensive array of NLP tasks. However, most existing works are limited to the textual modality, where both the model outputs and the feedback are in textual form. The recent surge in multi-modal data usage, including image, audio, and video modalities, presents enticing opportunities for expansion. These include the exploration of self-correction capabilities within multi-modal LLMs, the incorporation of visual feedback, and improving vision-language tasks through self-correction. 

A handful of pioneering studies have investigated this domain. For example, \textit{MaskGIT}~\cite{DBLP:conf/cvpr/ChangZJLF22} employed a self-refinement approach to image generation, where the model progressively refines the generated image conditioned on the previous generation. \citet{DBLP:conf/cvpr/KeLBHGLGCS19} utilized a self-correction strategy for vision-and-language navigation. FLIRT~\cite{mehrabi2023flirt} uses self-correction to iteratively generate and revise adversarial input prompts for text-to-image generation. However, despite these initial explorations, self-correction strategies are yet to be broadly adopted in multi-modal settings. A comprehensive understanding of how self-correction methods generalize across various modalities is crucial for improving their robustness and versatility.

\section{Conclusion}

In this paper, we present a comprehensive survey of self-correcting large language models with automated feedback. We broadly categorize and analyze various self-correction strategies, including training-time, generation-time, and post-hoc corrections. We also discuss the major application areas of self-correction, including correcting factual errors, enhancing reasoning abilities, and improving code generation, among others. Finally, we outline a number of potential future directions and associated challenges in this field. Our goal with this paper is to provide a comprehensive and useful resource for readers interested in the development of this rapidly evolving domain. To aid in this effort, we create a continually-updated reading list in a GitHub repository: \url{https://github.com/teacherpeterpan/self-correction-llm-papers}. 

% Directly manipulating and editing the model's parameters could open new horizons for self-correction in Large Language Models (LLMs). This approach, termed as "Model Editing," potentially offers a more targeted and efficient method for self-correction than traditional retraining methods.

% self-correction
% follows the paradigm of automatically detecting errors -> refine the error or improve the model 

% model editing
% alter the model's behavior towards a certain direction (target)

% difference: altering does not necessarily mean correction bad behaviours. out-of-date information, etc. 

% The goal of model edit is to adjust the model's behavior for examples within the editing scope while leaving its performance for out-of-scope examples unaltered. 

% The target scope is given. 

% However, techniques of model editing can be naturally incorporated for self-correction, in which a critic model is responsible of finding and locating diagnosing potential misbehavior of LLMs and model editing techniques are applied to 
% no existing works that combine these two areas. 

%\section*{Limitations}
%Required, may be limited in this kind of study.

%\section*{Ethics Statement}
%Required, may be limited in this kind of study.

\section*{Acknowledgements}
This work was supported by the National Science Foundation Award \#2048122. The views expressed are those of the authors and do not reflect the official policy or position of the US government. Thanks to Xinyuan Lu for assisting with the Github reading list repo. 

\bibliography{anthology,custom}
\bibliographystyle{acl_natbib}

%\appendix
%\section{Example Appendix}
%\label{sec:appendix}
%This is a section in the appendix.

\end{document}